\documentclass[letterpaper]{article} %
\usepackage{aaai24}  %
\usepackage{times}  %
\usepackage{helvet}  %
\usepackage{courier}  %
\usepackage[hyphens]{url}  %
\usepackage{graphicx} %
\usepackage{natbib}  %
\usepackage{caption} %
\usepackage{algorithm}
\usepackage{algorithmic}

\usepackage{multirow}
\usepackage{tabularx}
\usepackage{ragged2e}
\usepackage{subcaption}
\usepackage{adjustbox}
\usepackage{array}

\newcolumntype{R}[2]{%
    >{\adjustbox{angle=#1,lap=\width-(#2)}\bgroup}%
    l%
    <{\egroup}%
}
\newcommand*\rot{\multicolumn{1}{R{45}{1em}}}%

\title{Improving Knowledge Extraction from LLMs \\ for Task Learning through Agent Analysis}

\author{
    James R. Kirk,
    Robert E. Wray,
    Peter Lindes,
    John E. Laird
}
\affiliations{

    Center for Integrated Cognition at IQMRI \\
    Ann Arbor, MI 48105 USA \\
    \{james.kirk,robert.wray,peter.lindes,john.laird\}@cic.iqmri.org
}

\nocopyright 

\begin{document}
\maketitle

\begin{abstract}

Large language models (LLMs) offer significant promise as a knowledge source for task learning. 
Prompt engineering has been shown to be effective for eliciting knowledge from an LLM, but alone it is insufficient for acquiring relevant, situationally grounded knowledge for an embodied agent learning novel tasks.
We describe a cognitive-agent approach, STARS, that extends and complements prompt engineering, mitigating its limitations and thus enabling an agent to acquire new task knowledge matched to its native language capabilities, embodiment, environment, and user preferences. The STARS approach is to increase the response space of LLMs 
and deploy general strategies, embedded within the autonomous agent, to evaluate, repair, and select among candidate responses produced by the LLM.
We describe the approach and experiments that show how an agent, by retrieving and evaluating a breadth of responses from the LLM, can achieve $77-94\%$ task completion in one-shot learning without user oversight. The approach achieves $100\%$ task completion 
when human oversight (such as an indication of preference) 
is provided. Further, the type of oversight largely shifts from explicit, natural language instruction to simple confirmation/discomfirmation of high-quality responses that have been vetted by the agent before presentation to a user.
\end{abstract}

\section{Introduction}

Prompt engineering \cite{reynolds_2021}, along with in-context learning \cite{openai2023gpt4}, has been shown to be an effective strategy for extracting knowledge from a large language model (LLM).
However, embodied agents learning task knowledge (e.g., goals and actions) face far more stringent requirements. LLM responses must be:
\begin{enumerate}
    \item Interpretable by the agent's parsing capabilities. LLM responses must be understandable by the agent, meaning grammar and terminology are presented in a form that the agent can actually process.
    \item Situated to the agent's environment. Objects, features, and relations referenced in an LLM response must be perceivable and identifiable in the environment for the agent to ground the response successfully.
    \item Matched to agent's embodiment and affordances. An LLM, trained on a large corpus describing human activities, will (generally) generate responses conforming with human embodiment and affordances. Responses that do not consider an agent's often non-human embodiment (e.g., a single-armed robot) will often be infeasible for that agent to execute.
    \item Aligned with individual human preferences and values. Users will have individual expectations about how tasks should be performed and what constitutes appropriate outcomes in the current situation. Task success requires identifying and conforming to these preferences.
\end{enumerate}

The first three requirements are necessary for an embodied agent to use an LLM response to act in its world. We define responses that meet these requirements as \textit{viable}. The final requirement is necessary to achieve the task as a specific human user prefers. A response is \textit{situationally relevant} if it is viable \emph{and} matches the user's preferences.

To attempt to elicit viable responses from the LLM, we previously \cite{kirk2023integrating} employed a template-based prompting approach \cite[TBP;][]{olmo_gpt3--plan_2021_custom,kirk2022improving,reynolds_2021}. We developed prompt templates that included examples of desired task knowledge, instantiated them with context from the current task, and retrieved multiple responses (varying LLM temperature to generate different responses). Unfortunately, this TBP strategy produced responses that often violated one or more of the first three requirements.  Human feedback could be used to overcome these limitations, but required substantial input to correct responses (as well as to align them with agent needs and user preferences), making TBP impractical for an embodied agent. 

Motivated by these inadequacies, we present a novel strategy: Search Tree, Analyze and Repair, and Selection (STARS). Similar to ``agentic'' uses of LLMs \cite{sumers_cognitive_2023,park2023generative,Significant-Gravitas/Auto-GPT}, we employ the LLM as a component  within a larger system. Like self-consistency \cite{wang_self-consistency_2023_custom}, STARS generates a large space of responses from the LLM (multiple responses to a query). In contrast with the voting in self-consistency, the agent analyzes and evaluates each response for potential issues (e.g., mismatched embodiment, unknown words, ungrounded references). It attempts to repair problematic responses via targeted re-prompting of the LLM. To select among candidates, the agent queries the LLM for a ``preferred'' response.
The STARS agent can still solicit human feedback, but the primary purpose of oversight is to ensure that agent behavior (and learning) incorporates user preferences.

To evaluate STARS against TBP, we embed both methods within an existing embodied agent \cite{mohan_learning_2014_custom,mininger_expanding_2021,kirk_learning_2016}. This agent uses interactive task learning \cite[ITL;][]{laird_interactive_2017,gluck_interactive_2019_custom} to learn novel tasks via natural language instruction from a human user. Instead of querying a human for a goal description of the task (e.g., ``the goal is that the can is in the recycling bin''), the new agents (using TBP or STARS) access the LLM for that goal.

We compare STARS to TBP and also evaluate the individual components of STARS (i.e., Search Tree, Analysis \& Repair, Selection) in a simulated robotic environment. We assess both task completion rate and the amount of oversight needed to achieve 100\% task completion. We hypothesize STARS will eliminate the need to solicit human feedback for unviable responses, resulting in a much higher task completion rate (without oversight) and reducing how much oversight is required when human input is available. 

As we show below, over three different tasks, STARS achieves 77-94\% task completion without oversight (in comparison to 35-66\% with TBP). 
With oversight, STARS reduces the number of words needed from the user 
by 52-68\% (compared to TBP). Further, providing oversight is much simpler for the user. The user no longer needs to evaluate the viability of responses nor provide (many) goal descriptions; now, the user largely indicates preference, simply confirming or disconfirming from the LLM responses that the agent has determined to be viable. 
Finally, because the original ITL agent learns long-term task and subtask knowledge in one shot, this new agent also demonstrates one-shot performance: it achieves $100\%$ task completion when prompted to perform the same task in the future, without accessing the LLM or requiring further human input.

\section{Related Work}
Core features of our approach are 1) online task learning (no pre-training for domain or task), 2) the exploitation of multiple sources of knowledge, 3) proactive evaluation of LLM responses, and 4) one-shot task learning. We review related work in terms of these solution features.

Inner Monologue \cite{huang2022inner_ijcai} modifies its prompts based on feedback from the environment, agent, and user to elicit new responses when an action fails. Repair focuses on a single response at a time; STARS analyzes a set of responses to evaluate the result of using them, making evaluations and repairs before any response is selected and used. 
\citet{logeswaran_few-shot_2022_custom} plan sequences of subgoals from multiple LLM responses obtained from beam search (as in STARS) that does re-ranking based on feedback from the environment. 
SayCan \cite{ahn2022can_ijcai} uses an LLM and a trained set of low-level robot skills with short language descriptions for objects. The LLM is prompted multiple times for a high-level task to retrieve one low-level step at a time until a complete plan is found.
To obtain knowledge of low-level tasks, SayCan is trained on over 68K teleoperated demonstrations and human-rated simulations. STARS encodes properties for object classes (e.g., whether an object can be ``grabbed" by the robot) but requires no pre-training or prior exposure to the domain.

TidyBot \cite{wu2023tidybot} and TIDEE \cite{sarch2022tidee} address robotic problems similar to one of our experimental tasks (tidying a kitchen). They also account for human preferences. TidyBot tries to elicit human preferences by having the LLM summarize a few answers given by a human. TIDEE attempts to learn preferences by using ``commonsense priors'' learned previously by performing tasks in a ``training house.'' 
STARS does not depend on pre-training, but does elicit human preferences via NL dialogues.

PROGPROMPT \cite{singh2022progprompt} produces task plans by prompting an LLM with Python code that specifies the action primitives, objects, example tasks, and task name. The LLM returns a task plan in Python which includes assertions about states of the environment that are checked during execution, and recovery steps if an assertion fails. 
STARS retrieves NL descriptions of goals, rather than plans, and evaluates goals before they are used.

STARS attempts to verify LLM responses before attempting to achieve the goal indicated by a response. There are many approaches to verification of LLM knowledge, including 1) response sampling \cite{wang_self-consistency_2023_custom}, 2) use of other sources of knowledge  such as  planning \cite{valmeekam2023planning} or an LLM \cite{kim_language_2023_custom}, and 3) human feedback/annotation (TidyBot). 
Recursively Criticizes and Improves \cite[RCI;][] {kim_language_2023_custom} verifies LLM output by prompting the LLM again to identify (potential) issues. 
\citet{cobbe2021training} train a verifier to rank responses, while self-consistency \cite{wang_self-consistency_2023_custom} uses voting  to select an answer.
\citet{diao_active_2023_custom} combine all three of the above verification strategies by eliciting responses from an LLM, ranking them using an uncertainty metric (a source of knowledge other than the LLM), and then having humans annotate responses for further exploration.

While these efforts address similar challenges (or aspects of them), a unique aspect of STARS is the proactive analysis of many responses retrieved via prompting an LLM through embodied reasoning. The analysis enables the identification of known problems and targeted repairs. STARS also learns goal states for tasks, rather than action sequences to achieve tasks.
The STARS agent learns task knowledge in one shot, during performance, without prior training. When confronted with the same or similar tasks in the future, the agent can efficiently execute the task without the use of the LLM (or STARS). Encoding persistent task knowledge contrasts with in-context learning \cite{openai2023gpt4}.

\section{Prior Baseline: Template-based Prompting}
\begin{figure}
    \centering
    \includegraphics[width=0.95\linewidth]{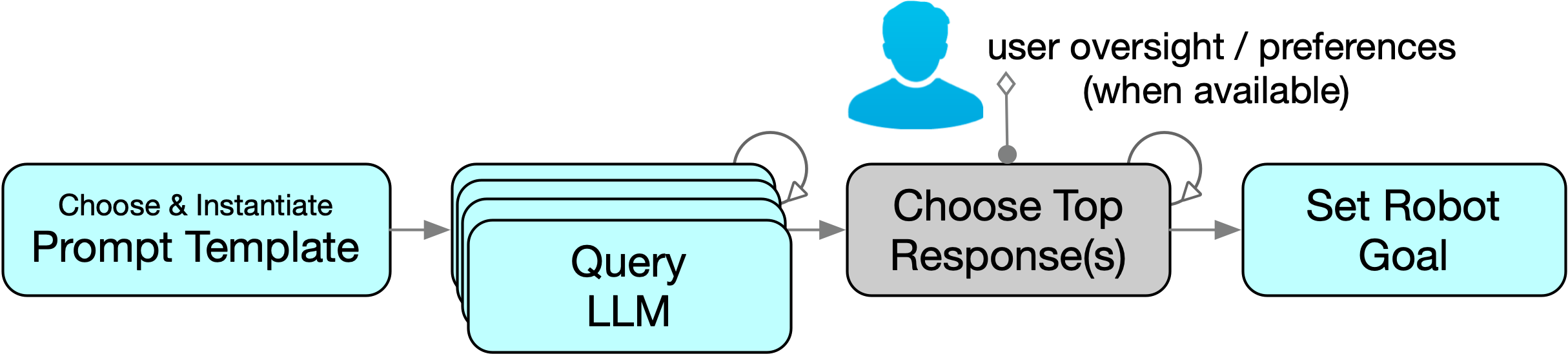}
    \caption{Baseline approach to elicitation of goal descriptions via template-based prompting (TBP).}
    \label{fig:prompt-engineering}
\end{figure}

The agent employs template-based prompting (TBP) to elicit responses from the LLM. Templates enable the agent to construct prompts using context from the task and environment and introduce prompt examples matched to the agent's capabilities and embodiment. Figure \ref{fig:prompt-engineering} outlines the baseline template-based prompting approach for generating task-goal descriptions (i.e., it replaces the NL-dialogue for ``Get goal description" in Figure~\ref{fig:retrospective-learning}). A prompt template is chosen and instantiated with relevant context, the LLM is queried (potentially soliciting multiple responses using varying temperatures), and response(s) are chosen for execution. In this baseline approach, choices are ranked by the mean log probabilities of tokens in each response.
Oversight is used to select an LLM response or to give a goal description when all LLM-generated choices are unacceptable. The agent uses the chosen response to attempt to perform the task and, if successful, learns a policy to execute the task in the future (see Figure~\ref{fig:retrospective-learning}). Few-shot examples in the prompt bias the LLM toward responses that are viable and relevant, matching the agent's NLP capabilities, desired semantic content (e.g., simple goal statements), and embodiment limitations \cite{kirk2022improving}.
This baseline approach learns the task in one shot but requires substantial user oversight to overcome errors \cite{kirk2023integrating}.

\section{The STARS Approach}
STARS extends the TBP baseline with three processes: retrieving a tree of LLM responses via beam search (ST: Search Tree), analyzing and repairing responses (AR: Analysis and Repair), and using the LLM to select a goal response from the candidates (S: Selection). 
After presenting each of these components of STARS, we describe the oversight strategy of soliciting user feedback. 

Figure \ref{fig:approach} outlines the process of the STARS approach (blue boxes are re-purposed elements from TBP; green boxes are new components of STARS). With STARS, the agent retrieves goal descriptions from the LLM (the rest of the task-learning process is the same). STARS ensures that the goal descriptions it retrieves from the LLM are viable for the agent. 
Acquiring goal knowledge is crucial to learning novel tasks, enabling an agent with planning capabilities to perform the new task. Goal learning enables greater flexibility than learning a sequence of actions because goal-state knowledge can transfer to other situations that require different action sequences to achieve the same goal.

\subsection{Search Tree (ST)}
In prior work with TBP (Figure~\ref{fig:prompt-engineering}), 
we increased the temperature parameter iteratively to retrieve multiple responses for the same prompt. This approach resulted in many duplicate responses and more responses that were not viable, deviating from targeted content and form. Similar to others \cite{logeswaran_few-shot_2022_custom,wang_self-consistency_2023_custom}, here we enable the agent to use a beam-search strategy to generate a breadth of high-probability responses from a single prompt.

\subsection{Analyze and Repair (AR)}
While many responses retrieved from the LLM are reasonable, they often fail to meet other requirements: being 
matched to the agent's embodiment, language capabilities, and situation. An agent that attempts to use a mismatched response will fail. Analysis and Repair detects and categorizes mismatches, drawing on the cognitive agent's knowledge and capabilities to identify problems, and then attempts to repair responses with identifiable mismatches.

\begin{figure}[t]
    \centering
    \includegraphics[width=1.0\columnwidth]{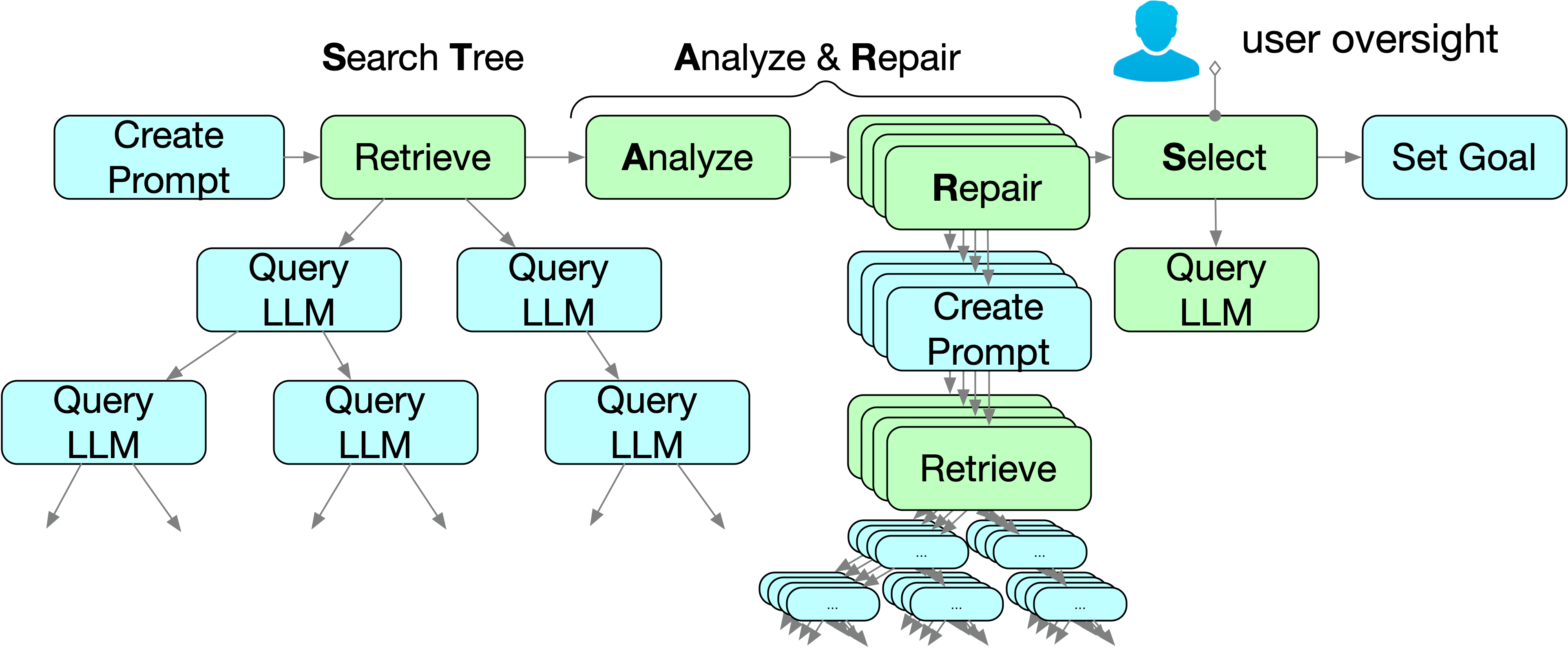}
    \caption{Summary of STARS approach.}
    \label{fig:approach}
\end{figure}

\begin{figure}[t]
    \centering
    \includegraphics[width=1.0\columnwidth]{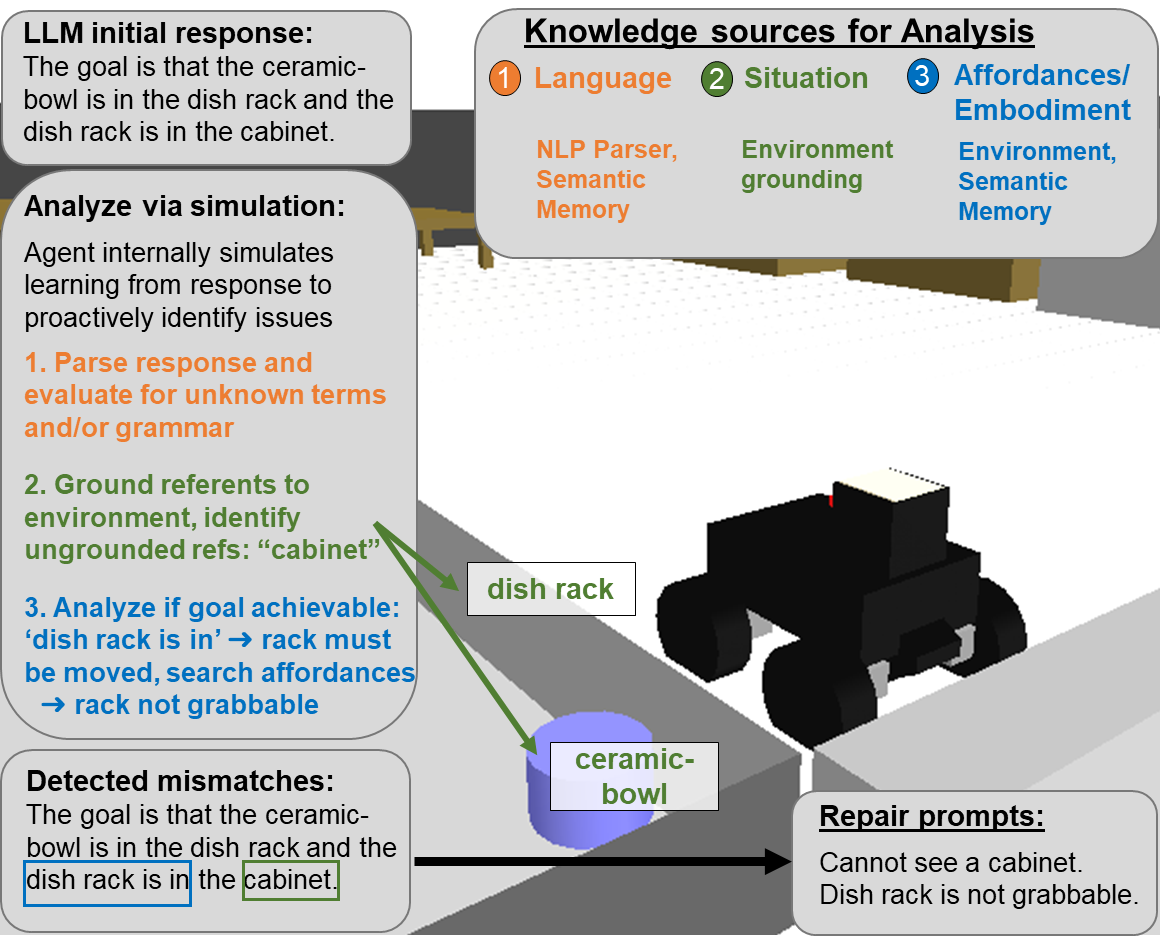}
    \caption{Agent analysis of mismatches via internal simulation}
    \label{fig:analysis}
\end{figure}

The overall process for Analysis and Repair is illustrated in 
Figure \ref{fig:analysis}. The agent performs a mental simulation of \textit{what would happen} if it attempted to use a response from the LLM, using the same knowledge of parsing and grounding it uses when performing the task. The analysis evaluates interpretability (orange: whether the agent can parse and interpret the language and terms), grounding (green: whether each referent in the response can be grounded to an object observable in the environment), and affordances (blue: whether the agent can achieve the actions on objects implied by clauses in the goal response).
The ``AR'' process currently addresses these three sources of mismatch: 

\begin{itemize}
    
    \item \textbf{Language}: The agent parses the response with its native NLP capabilities and examines the output. The language processor indicates if a sentence can be interpreted and identifies unknown words.

    \item \textbf{Situation}: To detect grounding issues, the agent evaluates the results of its language comprehension process. When a sentence contains a referring expression to an object, such as a cabinet, the agent's language processing identifies grounding candidates observable by the agent. Failure to ground a referent 
    indicates a mismatch with the current situation.

    \item \textbf{Embodiment and Affordance}: The agent detects embodiment and affordance mismatches using its knowledge of objects (semantic memory) and properties detected from perception (environment). E.g., when it processes a clause in a goal response  such as 
    ``the dish rack is in the cabinet,'' it evaluates if the object to be moved (``dish rack") has the property ``grabbable.''
    
\end{itemize}

Repair is coupled to these diagnostic mismatches detected during analysis. 
For each type of diagnosis, the agent constructs a new prompt using a repair template for that category of mismatch. The agent instantiates the template by appending the non-viable response with an instruction indicating the specific mismatch that occurred, e.g., ``No. Cannot see a cabinet.'' or ``No. Rack is not grabbable.''\footnote{A Technical Appendix provides complete examples of  prompts for repairs and selection:  \url{https://arxiv.org/abs/2306.06770}.} 
ST then uses this repair prompt to generate a new tree of responses.

\subsection{Selection (S)}
ST and AR are designed to generate viable candidate responses. However, the agent must select a single response to use. 
Rather than using mean log probability (as in TBP; Figure~\ref{fig:prompt-engineering}) or voting \cite[as in self-consistency][]{wang_self-consistency_2023_custom}, the new Selection strategy employs the LLM for choosing a response. The agent constructs a prompt with the candidates and asks which of a numbered list of candidate responses is the most reasonable goal given the task context.
The prompt solicits a single integer response from the LLM, indicating which response is the best.

\subsection{User Oversight (O)} \label{section:oversight}
The correct goal for some tasks depends on human preferences (e.g., some users prefer storing cereal in the cupboard, others, the pantry). The original ITL agent solicited all task knowledge from a human, which naturally captured this preference knowledge. STARS reduces user interaction while still ensuring capture of preference. Having the human in the loop also ensures correct learning. The agent solicits user feedback by asking if a retrieved goal is correct (yes/no) before using it (below). Selection determines which option to present. If the first response is rejected, Selection is repeated with the rejected option removed. If all responses are rejected, the user must provide the correct goal description.
\begin{quote}
\textbf{Agent}: For a mug in the dish rack is the goal that the mug is in the cupboard and the cupboard is closed?

\textbf{User}: Yes.
\end{quote}

\section{Experiment Design}
\label{sec:result}
In order to evaluate STARS, we first describe the embodied agent that incorporates STARS, an experimental design, and measures. In the next section, we present results for online learning of three different tasks: tidying the kitchen, storing groceries, and organizing an office. We evaluate how well STARS addresses the above requirements and also examine the relative impact of components of STARS. STARS learns descriptions of goal states, while systems such as SayCan, InnerMonologue, and TidyBot learn action sequences. We do not directly compare performance for these tasks against these systems because of their different learning targets.

\begin{figure}
    \centering
    \includegraphics[width=0.95\linewidth]{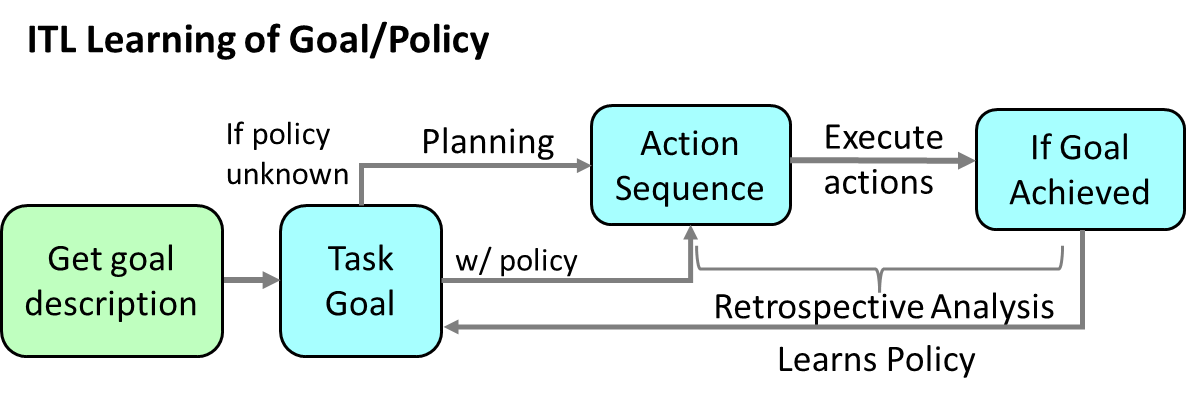}
    \caption{ITL process for learning goals and policy.}
    \label{fig:retrospective-learning}
\end{figure}
\textbf{Agent:} We embed STARS in an existing embodied ITL agent, replacing the human interaction that provided natural language descriptions of goals for tasks and subtasks.\footnote{Code for the ITL agent with STARS, simulator, and data analysis are available at \url{https://github.com/Center-for-Integrated-Cognition/STARS}.} The original agent learns a variety of diverse tasks (from puzzles to mobile patrol tasks) in many different physical (Fetch robot, mobile robot, and tabletop arm) and simulated (AI2Thor, April simulator) robotic domains \cite{mohan_acquiring_2012,mininger_expanding_2021,kirk_learning_2019}.

Figure~\ref{fig:retrospective-learning} depicts the ITL process for learning goals. The ITL agent can also learn new concepts, new actions (when planning knowledge is insufficient), and lower-level skills via instruction (not shown here). We focus on the goal-learning pipeline here because STARS exploits an LLM to learn goal descriptions (replacing the green box) without changing other aspects of the pipeline. The ITL learning process depended on substantial user input to provide interpretable and accurate descriptions of goals.
When a policy for achieving a goal is unknown, internal planning finds a sequence of actions that achieves the goal. A side effect of successful planning is that the agent learns long-term policy knowledge in one shot via the agent architecture's procedural learning mechanism. 
When the task arises in the future, that learned knowledge guides agent decision-making without planning or human interaction.

\textbf{Setting}: A simulated office and kitchen with a mobile robot created in the APRIL MAGIC simulator. The robot can move around the room, approach objects, and has a single arm that can grasp and manipulate all objects relevant to the task to be learned. For the ``tidy kitchen'' task (the largest task), the kitchen is populated with 35 objects that commonly exist in a kitchen (plates, condiments, utensils, etc.). Objects are initially distributed on a table, counter, and in the dish rack. For the ``store groceries'' task, 15 objects are contained in bags on the kitchen floor that must be stored (into the fridge, cupboard, or pantry). For the ``organize office'' task, 12 objects are distributed on a desk that must be cleared (into the drawer, bookshelf, trash, recycling bin, or filing cabinet). The three tasks contain 58 unique objects for which the agent needs to learn a goal.

\textbf{Simulation}: Although prior work with the ITL agent has used a physical robot, this experiment is done in simulation, which is sufficient for investigating the grounding of concepts and interpreting and learning from the descriptions provided by STARS.

\textbf{Learning Task}: For each experiment, the user presents the task (e.g., ``tidy kitchen'') and primary subtasks (e.g., clearing, storing, and unloading all the objects from the table, counter, and dish rack). For all tasks, task success is measured by the fraction of objects moved to a location consistent with user preferences. Also, when another object is manipulated to achieve a task (e.g., opening a refrigerator door to put away ketchup), it must be in its desired state for the task-success evaluation (e.g., the door must be closed). For the ``tidy kitchen'' task, four object types have multiple instances that must be treated differently based on their positions (e.g., a mug on the table must be put in the dishwasher or sink, but a mug in the dish rack must be put in the cupboard). 
Using the approach in Figure~\ref{fig:approach} (or a STARS variant as below), the agent acquires goal descriptions for each perceived object. It then uses the processing described in Figure~\ref{fig:retrospective-learning} to learn the goal and action policy, 
 enabling it to correctly process that object in the future without the LLM, planning, or oversight.

\begin{table}
    \centering
    \begin{tabularx}{\linewidth}{rXrrrrr}
\hline
\textit{Condition} & \textit{Description}  \\
\hline
TBP & \textbf{T}emplate-\textbf{B}ased \textbf{P}rompting (Baseline) \\
TBP+O & TBP with human \textbf{O}versight \\
ST & Beam \textbf{S}earch \textbf{T}ree  \\
STS & Beam search with LLM \textbf{S}election  \\
STAR & Beam search with  \textbf{A}nalysis (check viability) and  \textbf{R}epair \\
STARS & Search-tree, A\&R, LLM Selection \\
STARS+O  & STARS with human oversight.  \\
Trial \#2  & Task performance on second presentation after learning with STARS+O. \\
\hline        
    \end{tabularx}
    \caption{Definition of experimental conditions.}
    \label{tab:conditions}
\end{table}

\begin{table}[t]
    \centering
    \begin{tabularx}{\linewidth}{rrrrrrr}
\hline
\textit{Condition} & \multicolumn{1}{p{.75cm}}{Comp. ($\%$)} & \multicolumn{1}{p{.75cm}}{Goals \par retvd} & \multicolumn{1}{p{.75cm}}{Total \par tokens} & \multicolumn{1}{p{.75cm}}{\# \par instrct} & \multicolumn{1}{p{.75cm}}{\# \par words} \\
\hline
\multicolumn{6}{c}{\textbf{Tidy kitchen}} \\
\hline
TBP & 52.5 & 93 & 41407 & 14 & 76 \\
TBP+O & 100.0 & 89 & 42469 & 92 & 403 \\
ST & 50.0 & 243 & 56874 & 14 & 76 \\
STS & 40.0 & 247 & 66458 & 14 & 76 \\
STAR & 77.5 & 353 & 126086 & 14 & 76 \\
STARS & 77.5 & 368 & 139871 & 14 & 76 \\
STARS+O & 100.0 & 361 & 138096 & 65 & 127 \\
Trial \#2 & 100.0 & 0 & 0 & 1 & 2 \\
\hline
\multicolumn{6}{c}{\textbf{Store groceries}} \\
\hline
TBP & 66.7 & 39 & 17078 & 6 & 28 \\
TBP+O & 100.0 & 37 & 18689 & 29 & 92 \\
ST & 66.7 & 96 & 21518 & 6 & 28 \\
STS & 66.7 & 99 & 25690 & 6 & 28 \\
STAR & 77.8 & 170 & 57709 & 6 & 28 \\
STARS & 94.4 & 171 & 61808 & 6 & 28 \\
STARS+O & 100.0 & 177 & 64501 & 22 & 44 \\
Trial \#2 & 100.0 & 0 & 0 & 1 & 2 \\
\hline
\multicolumn{6}{c}{\textbf{Organize office}} \\
\hline
TBP & 35.7 & 34 & 12992 & 6 & 28 \\
TBP+O & 100.0 & 35 & 11662 & 41 & 184 \\
ST & 21.4 & 95 & 21082 & 6 & 28 \\
STS & 21.4 & 97 & 24717 & 6 & 28 \\
STAR & 64.3 & 204 & 75509 & 6 & 28 \\
STARS & 92.9 & 201 & 76056 & 6 & 28 \\
STARS+O & 100.0 & 206 & 77722 & 22 & 60 \\
Trial \#2 & 100.0 & 0 & 0 & 1 & 2 \\
\hline
    \end{tabularx}
    \caption{Summary of outcomes by condition for three tasks.}
    \label{tab:outcomes}
\end{table}

\textbf{Experimental conditions}: Experimental conditions are enumerated in Table~\ref{tab:conditions}. The TBP conditions are baselines for assessing the impact of the components of STARS. For all conditions, the LLM used is GPT-3 (for TBP, Search Tree, and Repair) and GPT-4 (for Selection).\footnote{GPT-4 does not currently expose logprobs, making it inapt for beam search. Selection does not use beam search and GPT-4 demonstrated better, more consistent results.} In all conditions, a user provides the initial task. In the Oversight conditions, the user reviews up to 5 responses. In non-oversight conditions, the choice of the goal is based on the highest mean log probability of candidates (ST and STAR) or the Selection strategy (STS and STARS).

\textbf{Measures:}
We assess conditions in three dimensions: performance, response quality, and cost. 
For performance, task completion rate (number of goal assertions achieved / total number of goal assertions) is the primary measure. For response quality, we evaluate how well responses align with requirements for situational relevance and viability, as well as reasonableness.
User effort is the largest factor impacting cost, but cannot be measured directly. To estimate effort, we use the number of interactions and words as well as the percentage of accepted goals. 
LLM costs are evaluated via tokens presented (prompts) and generated (responses).

\section{Experimental Results} 
The discussion of experimental results is organized around the three measures introduced above. Table~\ref{tab:outcomes} summarizes performance (task completion) and costs (tokens; oversight) for each condition for the three tasks. The Trial \#2 condition shows task performance after successful learning from STARS+O when given a second direction to perform the task; all tasks are completed successfully without further interaction beyond receiving the task (e.g., ``tidy kitchen'').\footnote{A video demonstration of STARS with a few objects is available at \url{https://youtu.be/bx7af5XAELY}.}

For each task we ran the STARS condition 10 times. Table~\ref{tab:variability} shows the mean values and standard deviation for task completion for each task. Due to the lack of variation between runs (attributable to the LLM and STARS) as well as experimental costs (GPT budget and the time to conduct each condition for all task experiments) we report results from one run for each condition (Table ~\ref{tab:outcomes}).  The overall variance for STARS is small and has a marginal effect on key outcomes (See Section D in the Technical Appendix for further exploration of variability in outcomes).

\begin{figure}[t]
    \centering
    \includegraphics[width=0.90\linewidth]{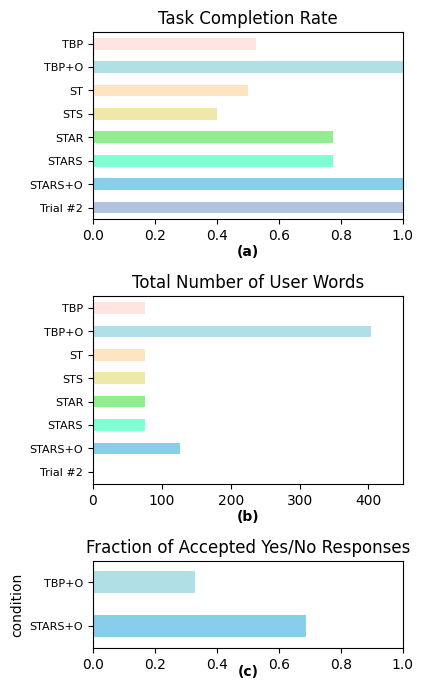}
    \caption{Performance and user cost measures for experimental conditions for the ``tidy kitchen'' task.}
    \label{fig:overall_results_typical}
\end{figure}

\textbf{Performance:}
Table~\ref{tab:outcomes} shows the task completion rates for all experimental conditions for the the three tasks. Figure~\ref{fig:overall_results_typical}(a) graphically compares task completion rates for the largest task: ``tidy kitchen.'' The baseline condition, TBP, achieves the experiment-defined targets (e.g., ``mug in the dishwasher") only 52.5\% (tidy kitchen), 66.7\% (store groceries), and 35.7\% (organize office) of the time. 
Adding Oversight to the baseline condition (TBP+O) results in 100\% task completion but vastly increases the number of required words (\ref{fig:overall_results_typical}b). Because many responses from the LLM are not viable and situationally relevant, the user must provide goal descriptions, resulting in many more words of instruction. Without oversight, STARS delivers a large gain in task completion, increasing to 77.5\% (tidy), 94.4\% (store), and 92.9\% (organize). Analysis and Repair (AR) prevents the agent from using unviable responses and increases the number of viable responses via repair. Search Tree (ST) alone results in no improvement but is a prerequisite for AR.

\begin{table}[tb]
    \centering
    \begin{tabular}{l|r|r|r}
    \hline
    Task: & Kitchen & Groceries & Office \\
    \hline
         Mean & 77.5 & 93.89 & 92.14 \\
         Std Dev. &   2.04 & 1.76  & 2.26 \\ 
         \hline
    \end{tabular}
    \caption{Variation in task completion rate for three tasks (STARS condition only).}
    \label{tab:variability}
\end{table}

\begin{figure}
    \centering
    \includegraphics[width=0.85\linewidth]{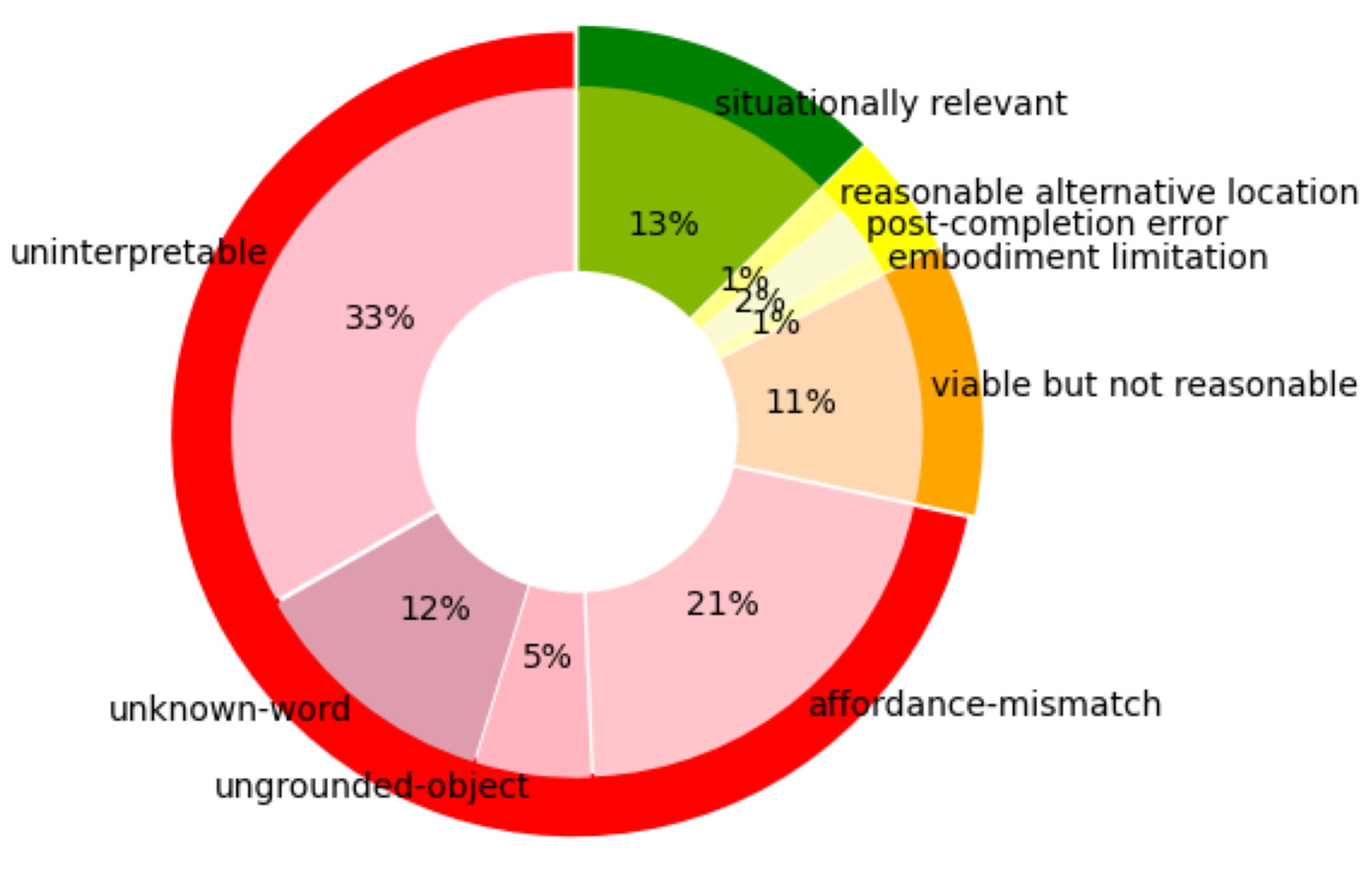}
    \caption{Categorization of responses retrieved from the LLM (STARS condition from ``tidy kitchen'' task).}
    \label{fig:response_categories}
\end{figure}

The task completion for ``tidy kitchen'' (77.5\%)  is significantly lower than for the other tasks using STARS. For the ``store groceries'' and ``organize office'' tasks, the addition of Selection (S) improved task completion, but did not for ``tidy kitchen.''
From detailed analysis, we determined that the agent lacks context specific to the tidy task. For instance, the agent (in this instantiation) lacks the ability to discriminate between a ``clean'' and ``dirty'' mug. In the ``tidy kitchen'' experiment, dishware on the table is assumed to be dirty (in terms of defining the target outcomes in the design), but the agent lacks this context. When such context is provided to the LLM (a variation we label STARS*),\footnote{Context provided to GPT-4 as a System prompt: ``Assume that dishware on the table or counter are dirty. Assume that bottles and cans are empty. Non-perishable food belongs in the pantry.''} Selection achieves 92.5\% task completion for ``tidy kitchen'' (without user oversight), comparable to the STARS task completion results for the other two tasks.
In the future, we will enable the user to provide this context directly.

With oversight, STARS task completion rises to 100\% for all tasks with much-reduced user input compared to TBP. This gain comes from shifting user input from providing goal descriptions (often needed in TBP) to confirming LLM-generated goal descriptions with yes/no responses (STARS+O). In addition, as highlighted in Figure~\ref{fig:overall_results_typical}(c), the greater precision of STARS in generating acceptable goal descriptions results in the user accepting a larger fraction of the goals in the oversight condition. The fraction of accepted goals increases from 33\% to 69\% (tidy kitchen), 62\% to 94\% (store groceries), and 18\% to 73\% (organize office).

\textbf{Quality of Responses:} Figure \ref{fig:response_categories} shows the percentage of different classifications of the responses retrieved from the LLM for STARS for tidying the kitchen.\footnote{Chart is representative of all conditions except TBP and Oversight; see appendix for each condition for all tasks.}
Responses are classified as unviable (red), viable but not reasonable (orange), reasonable (yellow), or situationally relevant (green). Further categorization identifies the type of mismatch for unviable responses (unknown word, ungrounded object, uninterpretable, affordance mismatch) and reasonable ones (reasonable alternative location, post-completion error, embodiment limitation). 
``Post-completion error" indicates a reasonable failure to close a door in situations where an object might not have a door. ``Embodiment limitation" captures when the robot places an object in a location that would otherwise be reasonable if its sensing were not limited.

Over 70\% of responses are not viable, leading to failure if the robot executed them; only 13\% are situationally relevant, meeting all four requirements. For storing groceries 58\% were not viable and 14\% were situationally relevant, and for organizing the office 85\% were not viable and only 5\% were situationally relevant. Thus, analysis of responses appears essential for reliable use of an LLM by an embodied agent to prevent the use of unviable goal descriptions. In the baseline (TBP) for tidying the kitchen, the agent retrieves at least one situationally relevant responses for only 15 of the 35 objects, while STARS results in 100\% of the objects having at least one situationally relevant response.\footnote{See appendix for graphical analysis of all conditions and tasks.}

\begin{figure}[t]
    \includegraphics[width=0.90\linewidth]{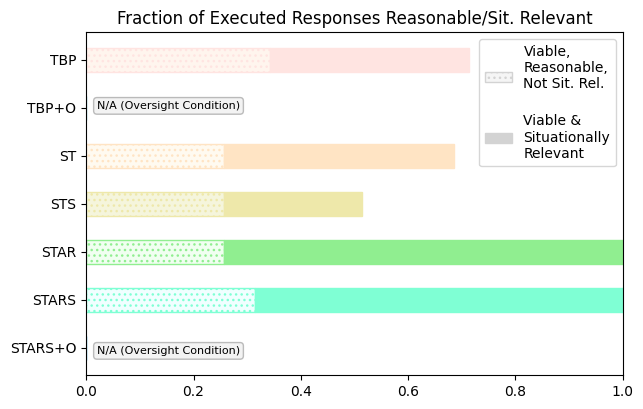}
    \caption{Fraction of responses used by the robot that are reasonable/sit. relevant for the ``tidy kitchen'' task.}
    \label{fig:objects_reasonable}
\end{figure}

Figure~\ref{fig:objects_reasonable} shows the quality of response by evaluating how frequently the robot receives a viable and (at least) reasonable response (situationally relevant for some user but not necessarily this one). For ``tidy kitchen,'' STARS (and STAR) results in 100\% of the used responses being at least reasonable. This indicates that STARS' 77.5\% task completion is close to the best it can achieve without oversight (or additional context). Human input is necessary to differentiate situationally relevant goals from reasonable ones.

\begin{figure}
    \centering
    \includegraphics[width=0.85\linewidth]{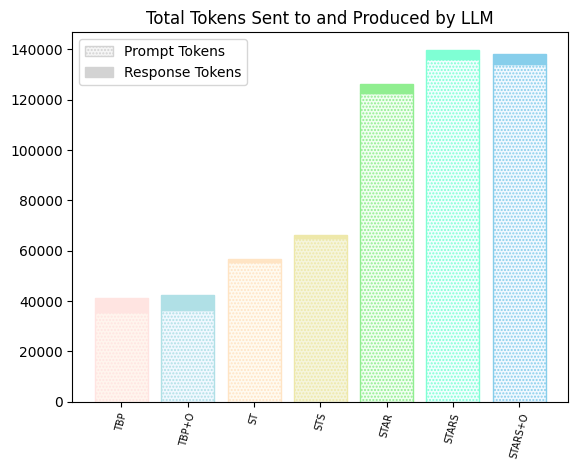}
    \caption{LLM tokens sent (hatched) and received (solid).}
    \label{fig:token_results_fig}
\end{figure}

\textbf{Cost:} Table~\ref{tab:outcomes} shows that oversight, in the form of instructions and words, is reduced by STARS (from 403 words to 127 for tidy kitchen, 92 to 44 words for store groceries, and 184 to 60 words for organize office). While the magnitude of the reduction is modest, the user now confirms a goal with a single word in comparison to supplying a complete goal description.
STARS+O also increases the precision of presented responses (Figure~\ref{fig:overall_results_typical}c); 69\% (kitchen), 94\% (grocery), and 73\% (office) of responses are accepted. Figure \ref{fig:token_results_fig} summarizes LLM tokens used for prompting and generation for ``tidy kitchen.''
For this task and the others, token cost increases substantially in Search Tree (ST) and Analysis and Repair (AR), because of the recursive beam search.

\section{Conclusion}
\label{sec:conclusion}
Using LLMs as the sole source of knowledge for an embodied agent is challenging due to the specific requirements that arise in operationalizing that knowledge.
STARS enables an agent to more effectively exploit an LLM, ensuring that the responses are viable (interpretable and grounded in the situation and agent capabilities). STARS shifts the role of the LLM from being the sole knowledge source to one source within a more comprehensive task-learning process \cite{kirk2023integrating}. It both addresses LLM limitations and takes advantage of the knowledge, reasoning, and online learning capabilities of cognitive agents.

While STARS provides significant improvements, further exploration and development are warranted. In particular, Selection does not provide a consistent improvement over the mean log prob choice strategy for ``tidy kitchen'' due to a lack of context. For future work, we will explore improvements to Selection, especially via the use of additional context that the agent can obtain from the user and (for some contexts) the LLM as briefly outlined here (STARS*). 

Finally, STARS also helps highlight the necessity of human oversight in the use of LLMs for agent task learning. Minimally, oversight ensures that an agent that uses an LLM is not led astray by the LLM, which can produce unsafe, biased, and unethical responses \cite{weidinger_ethical_2021}. Further, a human user will often be the only source of \textit{certain} knowledge of what goals and outcomes are appropriate for the task (Requirement 4). STARS, by ensuring that all candidates presented to the user are viable, simplifies and streamlines human oversight, reserving it for knowledge only a human can provide. 
This streamlining not only reduces the tedium of interaction (as suggested by the experimental results), it  also potentially allows users to better focus on alignment with their needs, goals, and values.

\section*{Ethical Statement}
This work uses large language models which can present ethical and social risks \cite{weidinger_ethical_2021} such as discrimination, exclusion, and toxicity or malicious uses. We consider each of these risks.

Because large language models are generative, depending on their corpus and training, they can produce language that reflects cultural biases, offensive stereotypes, derogatory usages, etc. In this work, where LLM queries are focused solely on producing goal descriptions for a particular task environment, we have not seen \textit{any} responses from GPT-3 that include such language.

In terms of exclusion, the specific tasks we have chosen do reflect (and mirror) cultural specificity to US/Western settings, in that the items in the kitchen and office (and the labels used to describe them) are both specific to the English language and typical of the objects one would find in a kitchen in a home or in an office environment. One of the long-term potential outcomes of this work (and ITL more generally) is that the agent is taught by human users in a particular setting, allowing the human to customize agent behavior to their specific setting (including its cultural context). Investigating whether this potential can be realized in the subject of future work.

Malicious use is also a potential risk, in that this research aims to enable human users to instruct agents to do their bidding. Users could theoretically instruct agents to cause direct harm to others, violate laws, etc. At this point in our research, this risk is minimal because implementations are confined to controlled, laboratory experiments. We are actively investigating in other work how an ITL agent can be both instructed while also following and conforming to both codified rules (like laws) and social norms to further mitigate the potential for malicious use.

\section*{Acknowledgements}
This work was supported by the Office of Naval Research, contract N00014-21-1-2369. The views and conclusions contained in this document are those of the authors and should not be interpreted as representing the official policies, either expressed or implied, of the Department of Defense or Office of Naval Research. The U.S. Government is authorized to reproduce and distribute reprints for Government purposes notwithstanding any copyright notation hereon.

\bibliography{zotero-transoar,verification-local}

\clearpage
\onecolumn
\appendix
\newcommand{\apptitle}[1]{\noindent{\centering\LARGE #1}}
\apptitle{\hspace{.38\linewidth} Technical Appendix}

\section{Objects in Experiments}
In this section we describe the objects used for the experiments in the paper.
Table \ref{tab:objects} shows the 35 objects used in the experiments for the ``tidy kitchen'' task, including their starting location in the kitchen and goal destination. All 35 objects listed have the property of ``grabbable" (can be picked up by the robot). The objects are distributed on the counter, table, and in the dish rack. The goal destinations of the objects are evenly distributed across the recycling bin, garbage, drawer, sink/dishwasher, cupboard, pantry, and refrigerator. 

\begin{table}[b!]
\centering
\begin{tabular}{l l l}
\hline
Object               & Location & Goal Destination \\ \hline
plastic-bottle       & table    & recycling bin     \\
soda-can             & table    & recycling bin     \\
coke-can             & counter  & recycling bin     \\
pepsi-can            & table    & recycling bin     \\
newspaper            & counter  & recycling bin     \\
apple-core           & counter  & garbage           \\
paper-plate          & table    & garbage           \\
plastic-fork         & table    & garbage           \\
plastic-cup          & table    & garbage           \\
paper-cup            & counter  & garbage           \\
paring-knife         & dish rack & drawer            \\
metal-fork           & dish rack & drawer            \\
\textbf{steak-knife}          & dish rack & drawer            \\
bottle-opener        & table    & drawer            \\
corkscrew            & counter  & drawer            \\
\textbf{ceramic-plate}        & table    & sink/dishwasher   \\
plate                & counter  & sink/dishwasher   \\
\textbf{glass-tumbler}        & table    & sink/dishwasher   \\
\textbf{steak-knife}          & counter  & sink/dishwasher   \\
\textbf{mug}                  & counter  & sink/dishwasher   \\
\textbf{mug}                  & dish rack & cupboard          \\
\textbf{glass-tumbler}        & dish rack & cupboard          \\
\textbf{ceramic-plate}        & dish rack & cupboard          \\
ceramic-bowl         & dish rack & cupboard          \\
coffee-grinder       & counter  & cupboard          \\
cereal-box           & table    & pantry            \\
box-of-aluminum-foil & counter  & pantry            \\
pop-tart-box         & table    & pantry            \\
granola-bars         & counter  & pantry            \\
crackers             & counter  & pantry            \\
milk                 & table    & refrigerator      \\
half-and-half        & counter  & refrigerator      \\
ketchup              & table    & refrigerator      \\
jar-of-salsa         & counter  & refrigerator      \\
apple-juice          & table    & refrigerator      \\ \hline
\end{tabular}
\caption{Objects used in ``tidy kitchen'' experiments with their starting locations and goal destinations.}
\label{tab:objects}
\end{table}

There are four duplicate object pairs (highlighted in bold), but due to their differing locations the goal state for each pair of duplicates is different (e.g., the steak knife on the table should be put in the dishwasher, the steak knife in the dish rack should be put in the drawer). For the goal destinations as designed in these experiments, dishes on the table or counter are treated as being dirty (reflecting the preferences of the user). However, some objects on the table must be treated differently. For example, the bottle-opener and cork screw have the goal of being placed directly into a drawer (as these objects are not typically washed after use). Using multiple instances of the same object type and having various different destinations from the same initial location were included in the design to 1) result in more challenging task to learn overall, and 2) evaluate how the LLM reacted to the different contexts.

\begin{table}
\centering
\begin{tabular}{l l l}
\hline
Object               & Location & Goal Destination \\ \hline
plastic-cups         & bag   & cupboard          \\
paper-plates         & bag   & cupboard          \\
flour                & bag   & pantry            \\
boxed-pasta          & bag   & pantry            \\
can-of-beans         & bag   & pantry            \\
granola              & bag   & pantry            \\
chips                & bag   & pantry            \\
yogurt               & bag   & refrigerator      \\
cream                & bag   & refrigerator      \\
hummus               & bag   & refrigerator      \\
apple-cider          & bag   & refrigerator      \\
cheese               & bag   & refrigerator      \\
orange-juice         & bag   & refrigerator      \\
eggs                 & bag   & refrigerator      \\
butter               & bag   & refrigerator      \\ \hline
\end{tabular}
\caption{Objects used in ``store groceries'' experiments with their starting locations and goal destinations.}
\label{tab:objects_groceries}
\end{table}

Table \ref{tab:objects_groceries} shows the 15 objects used in the experiments for the ``store groceries'' task, including locations and goal destinations. As before, all 15 objects listed have the property of ``grabbable" (can be picked up by the robot). The objects are distributed into three bags on the kitchen floor.

Table \ref{tab:kitchen-objects} shows the 11 appliances and furniture in the simulated kitchen that serve as the locations and destinations for objects in the ``tidy kitchen'' and ``store groceries'' experiments. It specifies properties of the objects that relate to what actions can be performed on them (affordances), including surface (objects can be placed on it) and receptacle (objects can be placed in it). It also lists if the objects have the affordance of openable/closeable. Finally it lists the goal state of the objects in the experiment design (e.g., that the ones that can be closed must be closed).

\begin{table}
\centering
\begin{tabular}{l l l}
\hline
Object        & Properties                    & Goal State \\ \hline
Table         & surface                       & N/A        \\
Counter       & surface                       & N/A        \\
Dish Rack     & receptacle                    & N/A        \\
Garbage       & receptacle                    & N/A        \\
Recycling bin & receptacle                    & N/A        \\
Pantry        & receptacle, openable/closeable & closed     \\
Cupboard      & receptacle, openable/closeable & closed     \\
Refrigerator  & receptacle, openable/closeable & closed     \\
Dishwasher    & receptacle, openable/closeable & closed     \\
Drawer        & receptacle, openable/closeable & closed     \\
Sink          & receptacle                    & N/A        \\ \hline
\end{tabular}
\caption{Appliance and furniture objects present in the simulated kitchen for experiments with their properties and goal states.}
\label{tab:kitchen-objects}
\end{table}

\begin{table}
\centering
\begin{tabular}{l l l}
\hline
Object                & Location & Goal Destination \\ \hline
folder                & bag   & filing cabinet            \\
file                  & bag   & filing cabinet            \\
paper-coffee-cup      & bag   & garbage            \\
tissue                & bag   & garbage            \\
plastic-water-bottle  & bag   & recycling bin      \\
sprite-can            & bag   & recycling bin      \\
dictionary            & bag   & bookshelf      \\
novel                 & bag   & bookshelf      \\
book                  & bag   & bookshelf      \\
stapler               & bag   & drawer      \\
pencil                & bag   & drawer      \\
pen                   & bag   & drawer      \\ \hline
\end{tabular}
\caption{Objects used in ``organize office'' experiments with their starting locations and goal destinations.}
\label{tab:objects_office}
\end{table}

Table~\ref{tab:objects_office} shows the 12 objects used in the experiments for the ``organize office'' task, including locations and goal destinations. As before, all 12 objects listed have the property of ``grabbable" (can be picked up by the robot). The objects are all on the desk in the office.
Table \ref{tab:office-objects} shows the 7 furniture objects in the simulated office that serve as the locations and destinations for objects in the ``organize office'' experiments. It specifies properties of the objects that relate to what actions can be performed on them (affordances), including surface (objects can be placed on it) and receptacle (objects can be placed in it). As before, iut lists if the objects have the affordance of openable/closeable. Finally it lists the goal state of the objects in the experiment design (e.g., that the ones that can be closed must be closed).

\begin{table}
\centering
\begin{tabular}{l l l}
\hline
Object          & Properties                     & Goal State \\ \hline
Desk            & surface                        & N/A        \\
Chair           & surface                        & N/A        \\
Filing cabinet  & receptacle, openable/closeable & closed        \\
Bookshelf       & receptacle                     & N/A        \\
Garbage         & receptacle                     & N/A        \\
Recycling bin   & receptacle                     & N/A        \\
Drawer          & receptacle, openable/closeable & closed    \\ \hline
\end{tabular}
\caption{Furniture objects present in the simulated office for experiments with their properties and goal states.}
\label{tab:office-objects}
\end{table}

\section{Step-by-Step Example of Goal Elicitation/Learning Process}
In this section we describe, with more detail and a running example drawn from the experiments, the complete learning process from the ``tidy kitchen'' task using the STARS strategy to retrieve a breadth of responses from the LLM (GPT-3), analyze and repair responses, and select from the candidate options (GPT-4). We focus on a single object and the STARS+Oversight condition with no ablations.

\begin{figure}[thb]
    \centering
    \includegraphics[width=0.60\linewidth]{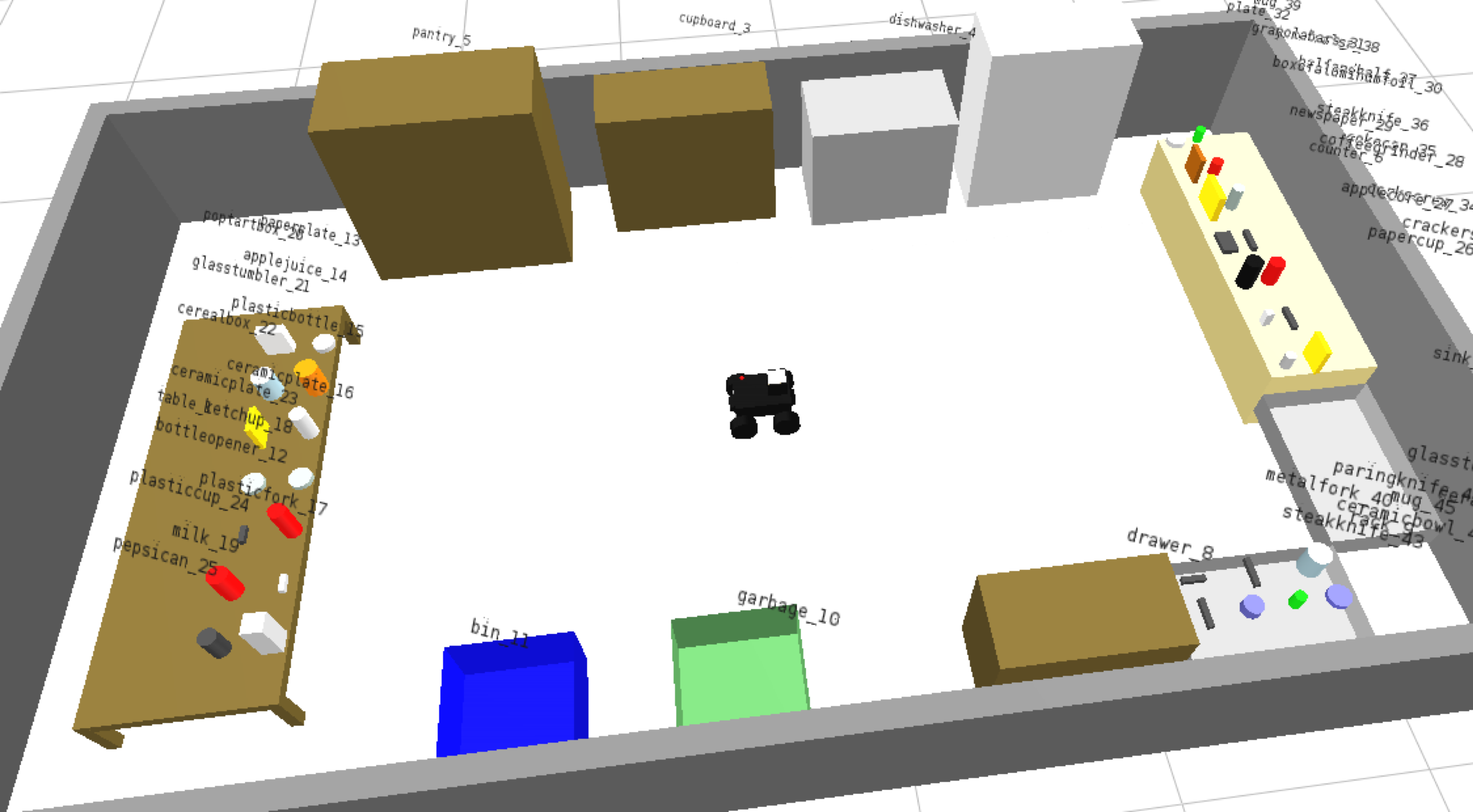}
    \caption{Setting for learning the ``tidy kitchen" task.}
    \label{fig:initial_conditions}
\end{figure}

Figure ~\ref{fig:initial_conditions} shows the simulation of the kitchen that is filled with objects that need to be tidied. We explore task learning in the context of a simulated office kitchen environment with a mobile robot that is capable of grabbing, moving, and interacting with objects.  First we describe initial interaction with the user, where the user gives the task and subtasks to learn, followed by the template-based prompting strategy that is the baseline for this approach and is used to select and instantiate a template for learning task knowledge. Finally we described the component strategies of STARS.

\subsection{Initial interaction, ITL agent}
The dialogue used to initiate teaching of the ``tidy kitchen'' task is listed below. The human provides the task to be learned, followed by subtasks to achieve clearing, storing, and unloading all the objects on the table, counter, and dish rack respectively. 

\begin{quote}
\textbf{Instructor}: Tidy kitchen. \\
\textbf{Robot}: What is the next goal or subtask of tidy? \\
\textbf{Instructor}: Repeat the following tasks while an object is on the table. \\
\textbf{Robot}: What is the next goal or subtask of tidy? \\
\textbf{Instructor}: Clear an object that is on the table. \\
\textbf{Robot}: What is the next goal or subtask of clear?
\end{quote}

The robot then looks for objects that it observes on the table. After selecting an object, the agent attempts to clear the object. If it does not know a goal for this object, the robot initiates the STARS learning processing to acquire a goal description, starting by constructing a prompt for the goal using the context of the task and the object. Once all objects have been cleared from the table, part of the above dialog will be repeated for storing objects that are on the counter, and again for unloading objects from the dish rack.

As an example going forward we will use a mug in the dish rack that the robot observes while learning to unload all the objects in the dish rack. The data is taken from the experiment for the STARS+Oversight condition.

\subsection{Prompt Construction}

Using template-based prompting, the robot selects a template for learning a task goal, which includes two prompt examples (from other tasks), and instantiates the prompt template with the relevant task context, the overall task ``tidy kitchen,'' the location of the robot ``in the kitchen,'' and the object observed ``mug in the dish rack.'' The initial prompt for the mug in the dish rack constructed by the robot is shown below.

This work focuses specifically on retrieving goal knowledge, which enables the robot to search for the steps to achieve the goal. Prior work has shown that, using this strategy, a robot can retrieve action knowledge with 97\% accuracy without additional interaction or evaluation by the robot, and therefore does not require the additional strategies proposed in this paper to find the actions needed to achieve a valid goal.

\begin{center}
\textbf{\emph{Agent-created prompt:}}
\end{center}
\begin{verbatim}
(EXAMPLES)(TASK)Task name: store object. Task context: I am in mailroom.
Aware of package of office supplies, package is in mailroom.
(RESULT)The goal is that the package is in the closet
and the closet is closed.(END RESULT)
Response: Ok.
Steps:
1. Open closet
2. Pick up package of office supplies
3. Put package into closet
4. Close closet
(END TASK)
(TASK)Task name: deliver package. Task context: I am in mailroom.
Aware of package addressed to Gary, package is in mailroom.
(RESULT)The goal is that the package is in Gary's office.(END RESULT)
Response: Ok.
Steps:
1. Pick up package addressed to Gary
2. Go to Gary's office
3. Put package onto desk in Gary's office
(END TASK)
(END EXAMPLES)
(TASK)Task name: tidy kitchen. Task context: I am in kitchen.
Aware of mug in dish rack.
(RESULT)
\end{verbatim}

\subsection{Search Tree}
To insure a range of responses to choose from, the robot uses beam search to retrieve a tree of responses for a single prompt. For any token with a log probability under 90\%, new completions will be generated for alternative tokens that are above 5\%. The logprobs GPT setting is set to 5, so four alternative responses will be retrieved from the LLM.
This process is recursive (up to a recursion depth of 3). To further limit the recursion and the number of responses generated, we also limited a second recursion of responses to those where the total response generated so far has a mean log probability above 85\%. These thresholds were selected after some pilot experiments, and are not tuned to the objects in the experiment data set. Lowering the thresholds results in a much larger space of responses retrieved.

First the temperature 0 response for the prompt is retrieved. From the above prompt for the mug the temperature 0 response is:
\begin{center}
\textbf{\emph{LLM response:}}
\end{center}
\begin{verbatim}
The goal is that the mug is in the dishwasher and the dishwasher is 
turned on
\end{verbatim}

For a deeper analysis of the beam search, the tokens of the response to the above prompt are listed below.

\newpage
\begin{center}
\textbf{\emph{Tokens in response:}}
\end{center}
\begin{verbatim}
[`The', ` goal', ` is', ` that', ` the', ` mug', ` is', ` in', ` the',
    ` dish', `washer', ` and', ` the', ` dish', `washer', ` is', ` turned',
    ` on']
\end{verbatim}

Of these tokens only ` dish,' `washer,' ` and,' and ` turned' have log probabilities below the threshold of 90\%. The log probabilities for each are shown in Table~\ref{tab:token-initial}, alongside the probabilities for each alternative potential response. Only tokens above the 5\% threshold (highlighted in bold) will be expanded in the beam search.

\begin{table}
\centering
\begin{tabular}{lllll}
\hline
Initial token    & \multicolumn{4}{c}{Alternative tokens}                                                                    \\ \hline
` dish' (0.483)  & \textbf{`   cup' (0.265)} & \textbf{`   cabinet' (0.213)} & `   sink' (0.0206)        & `   mug' (0.0088) \\
`washer' (0.793) & \textbf{` rack' (0.1658)} & ` cabinet' (0.0191)           & ` dr' (0.0158)            & ` cup' (0.00279)  \\
`   and' (0.881) & \textbf{`.('( 0.114)}     & `.' (0.00209)                 & `."'(0.00002582)          &                   \\
`turned' (0.536) & \textbf{`closed' (0.176)} & \textbf{`on'(0.1479)}         & \textbf{`started'(0.056)} & `full' (0.0380)  \\ \hline
\end{tabular}
\caption{Alternative tokens for tokens in the initial response under the threshold for beam search.}
\label{tab:token-initial}
\end{table}

\begin{center}
\textbf{\emph{Prompt for first level of recursion:}}
\end{center}
\begin{verbatim}
(EXAMPLES)(TASK)Task name: store object. Task context: I am in mailroom.
Aware of package of office supplies, package is in mailroom.
(RESULT)The goal is that the package is in the closet
and the closet is closed.(END RESULT)
Response: Ok.
Steps:
1. Open closet
2. Pick up package of office supplies
3. Put package into closet
4. Close closet
(END TASK)
(TASK)Task name: deliver package. Task context: I am in mailroom.
Aware of package addressed to Gary, package is in mailroom.
(RESULT)The goal is that the package is in Gary's office.(END RESULT)
Response: Ok.
Steps:
1. Pick up package addressed to Gary
2. Go to Gary's office
3. Put package onto desk in Gary's office
(END TASK)
(END EXAMPLES)
(TASK)Task name: tidy kitchen. Task context: I am in kitchen.
Aware of mug in dish rack.
(RESULT)The goal is that the mug is in the cup
\end{verbatim}
\par
\par

The tokens of the response to this prompt are listed below.
\begin{center}
\textbf{\emph{LLM response tokens for first recursion:}}
\end{center}
\begin{verbatim}
['board', ' and', ' the', ' cup', 'board', ' is', ' closed']
\end{verbatim}

Again the relative probabilities of the tokens are examined to continue the beam search. As before, the alternative tokens for tokens below 90\% probability are shown in Table~\ref{tab:token-first-recursion}, and only tokens above the 5\% threshold (highlighted in bold) will be expanded in the beam search.

\begin{table}[b]
\centering
\begin{tabular}{lllll}
\hline
Initial token    & \multicolumn{4}{c}{Alternative tokens}                                                                    \\ \hline
`  and' (0.8779)  & \textbf{`    .(' (0.1190)} & `         .' (0.0010) & `  above' (0.0002) \\
`   cup' (0.810) & \textbf{` dish' (0.1864)} & ` kitchen' (0.0009)           & `door' (0.0009)            & `counter' (0.0006)  \\  \hline
\end{tabular}
\caption{Alternative tokens for tokens under the threshold in the first recursion response.}
\label{tab:token-first-recursion}
\end{table}

When Search Tree encounters an alternative token that contains a period, indicating the end of the sentence, such as for `and' above, it returns that completion as a response: ``The goal is that the mug is in the cupboard.'' Search Tree then continues the beam search recursion by generating a completion where `dish' is used in place of `cup' as shown in the prompt below.

\begin{center}
\textbf{\emph{Prompt for second level of recursion:}}
\end{center}
\begin{verbatim}
(EXAMPLES)(TASK)Task name: store object. Task context: I am in mailroom.
Aware of package of office supplies, package is in mailroom.
(RESULT)The goal is that the package is in the closet
and the closet is closed.(END RESULT)
Response: Ok.
Steps:
1. Open closet
2. Pick up package of office supplies
3. Put package into closet
4. Close closet
(END TASK)
(TASK)Task name: deliver package. Task context: I am in mailroom.
Aware of package addressed to Gary, package is in mailroom.
(RESULT)The goal is that the package is in Gary's office.(END RESULT)
Response: Ok.
Steps:
1. Pick up package addressed to Gary
2. Go to Gary's office
3. Put package onto desk in Gary's office
(END TASK)
(END EXAMPLES)
(TASK)Task name: tidy kitchen. Task context: I am in kitchen.
Aware of mug in dish rack.
(RESULT)The goal is that the mug is in the cupboard and the dish
\end{verbatim}

The LLM responds with another sequence of tokens:

\begin{center}
\textbf{\emph{Response for second level of recursion:}}
\end{center}
\begin{verbatim}
[' rack', ' is', ' empty']
\end{verbatim}
\par
\par

No further recursion is done beyond this point. A similar process is performed for the other branches of the response `tree.' After expanding the entire tree to this level, the final set of responses retrieved using Search Tree for a mug in the dish rack are sent to the robot for analysis. These responses with their probabilities are listed below.

\begin{center}
\textbf{\emph{Final list of goals produced by tree search:}}
\end{center}
\begin{verbatim}
the goal is that the mug is in the cabinet and the cabinet is closed (0.937)
the goal is that the mug is in the cupboard and the cupboard is closed 
    (0.935)
the goal is that the mug is in the dishwasher and the dishwasher is turned 
    on (0.925)
the goal is that the mug is in the dishwasher and the dishwasher is closed
    (0.899)
the goal is that the mug is in the cupboard and the dish rack is empty
    (0.898)
the goal is that the mug is in the dishwasher and the dishwasher is on
    (0.897)
the goal is that the mug is in the dishwasher and the dishwasher is started
    (0.8919)
the goal is that the mug is in the dish rack and the dish rack is empty
    (0.881)
the goal is that the mug is in the dish rack and the dish rack is tidy
    (0.870)
the goal is that the mug is in the dish rack and the dish rack is clean
    (0.865)
the goal is that the mug is in the dishwasher (0.8618)
the goal is that the mug is in the cupboard (0.86128)
the goal is that the mug is in the dish rack and the dish rack is in the
    cupboard (0.860)
\end{verbatim}

\subsection{(Agent) Analysis}

Once the Search Tree process has retrieved a set of high probability responses from the LLM, STARS continues by analyzing each of the candidate responses to detect mismatches and determine which ones are viable for the robot. Each candidate is analyzed to determine if it matches robot's NLP capabilities, embodiment, affordances, and current environment. 

This analysis is performed through internal simulation, where the robot simulates learning from the response to proactively identify mismatches. The robot's language processor indicates if a sentence can be interpretted and identifies unknown words. It evaluates the results of the language comprehension grounding process to identify any referents in the response that could not be grounded to objects in the environment observable by the robot. Finally the robot detects affordance and embodiment mismatches using it knowledge of objects (from semantic memory) and properties of objects (detected through perception of the env.) by evaluating if the clauses in the response are achievable given its knowledge of affordances.

The analysis categorizes responses as viable if they contain no mismatches, and for responses with mismatches identifies the category of mismatch and the specific issue. The viable goals for the mug are listed below.
\begin{center}
\textbf{\emph{Agent analysis determines the following are viable:}}
\end{center}
\begin{verbatim}
the goal is that the mug is in the cupboard and the cupboard is closed
the goal is that the mug is in the dishwasher and the dishwasher is closed
the goal is that the mug is in the dishwasher
the goal is that the mug is in the cupboard
\end{verbatim}

The goal responses that the robot determine are unviable are listed below, grouped by the type of mismatch.
\begin{center}
\textbf{\emph{Uninterpretable responses (Language mismatch):}}
\end{center}
\begin{verbatim}
the goal is that the mug is in the dishwasher and the dishwasher is turned
   on
the goal is that the mug is in the dishwasher and the dishwasher is on
the goal is that the mug is in the dish rack and the dish rack is tidy
the goal is that the mug is in the dish rack and the dish rack is clean
\end{verbatim}

In these cases the robot was not able to interpret these responses.

\begin{center}
\textbf{\emph{Responses with unknown terms (Language mismatch):}}
\end{center}
\begin{verbatim}
the goal is that the mug is in the dishwasher and the dishwasher is
    started (Unknown word started)
\end{verbatim}

The robot does not have a definition of `started' and identifies it as an unknown word.

\begin{center}
\textbf{\emph{Responses with ungrounded objects (Situation mismatch):}}
\end{center}
\begin{verbatim}
the goal is that the mug is in the cabinet and the cabinet is closed,
    (Ungrounded object cabinet)
\end{verbatim}

There is no cabinet in the kitchen that the robot can observe, so it fails to ground the referent of cabinet to an object.

\begin{center}
\textbf{\emph{Responses with an affordance mismatch (Embodiment/affordance mismatch):}}
\end{center}
\begin{verbatim}
the goal is that the mug is in the cupboard and the dish rack is empty
    (affordanch mismatch: rack cannot have property empty)
the goal is that the mug is in the dish rack and the dish rack is empty
    (affordance mismatch: rack cannot have property empty)
the goal is that the mug is in the dish rack and the dish rack is in the
    cupboard(affordance mismatch: rack is not grabbable)
\end{verbatim}

For the affordance mismatches, the robot detects an affordance violation for the dish rack being empty because its affordance knowledge for empty relates to objects that can be filled with a liquid (e.g. a water pitcher) and it does not have the fillable affordance for the dish rack. The dish rack is also not an object that the robot is capable of grabbing or moving, so it identifies an affordance mismatch that the rack is not grabbable.

\subsection{Repair}

Given the results of Analysis, the Repair strategy of STARS attempts to repair the detected mismatches by prompting the LLM again. It will attempt to repair three types of mismatches: ungrounded objects, unknown words, and affordance mismatches. For each type of mismatch the robot has a prompt template that it can instantiate that contains an example of repairing that type of mismatch (for another task). Otherwise the prompt template is the same as was used for the initial prompt (as seen in ST). The offending, mismatched responses is appended onto the prompt, followed by a response from the robot indicate the mismatch to repair.

Below we continue the learning process for the mug in the dish rack, by showing the repairs performed on the responses for each of the types of mismatch.

\begin{center}
\textbf{\emph{Repairing an ungrounded object:}}
\end{center}

The first response the robot tries to repair is the response with an ungrounded object, cabinet, that the robot could not perceive in its environment.

\begin{verbatim}
the goal is that the mug is in the cabinet and the cabinet is closed
\end{verbatim}

The robot selects a prompt template for repairing ungrounded object that includes an example of the repair. This prompt example, for an ungrounded shelf, can be seen at the beginning of the prompt below. The prompt is instantiated as before, but now with the mismatched response appended followed by the response from the robot indicating the mismatch to repair: ``No. Cannot see a cabinet.''

\begin{center}
\textbf{\emph{Prompt:}}
\end{center}
\begin{verbatim}
(EXAMPLES)(TASK)Task name: store object. Task context: I am in mailroom.
Aware of package of office supplies, package is in mailroom.
(RESULT)The goal is that the package is on the shelf.(END RESULT)
Response: No. Cannot see a shelf.
(RESULT)The goal is that the package is in the closet.(END RESULT)
Response: Ok.
Steps:
1. Pick up package
2. Put package into closet
(END TASK)
(TASK)Task name: deliver package. Task context: I am in mailroom.
Aware of package addressed to Gary, package is in mailroom.
(RESULT)The goal is that the package is in Gary's office.(END RESULT)
Response: Ok.
Steps:
1. Pick up package addressed to Gary
2. Go to Gary's office
3. Put package onto desk in Gary's office
(END TASK)
(END EXAMPLES)
(TASK)Task name: tidy kitchen. Task context: I am in kitchen.
Aware of mug in dish rack.
(RESULT)the goal is that the mug is in the cabinet and the cabinet
is closed(END RESULT)
Response: No. Cannot see a cabinet.
(RESULT)
\end{verbatim}

This initial temperature 0 response from this prompt is listed below.

\begin{center}
\textbf{\emph{Repair LLM Response:}}
\end{center}
\begin{verbatim}
['the', ' goal', ' is', ' that', ' the', ' mug', ' is', ' in', ' the', 
    ' sink', ' and', ' the', ' sink', ' is', ' full', ' of', ' water']
\end{verbatim}

STARS doesn't just retrieve a single response it uses the beam search strategy from Search Tree to retrieve a set of responses to the repair as before. We won't step through the process again, as it is the same as before. The final responses generated from this repair prompt are shown below. Some of them are duplicates with responses already generated. Note that none of these responses refer to a cabinet anymore.

\begin{center}
\textbf{\emph{Final output for repair of ungrounded cabinet:}}
\end{center}
\begin{verbatim}
the goal is that the mug is in the drawer and the drawer is closed
the goal is that the mug is in the sink and the sink is full of water
the goal is that the mug is in the dish rack and the dish rack is empty
    (duplicate)
the goal is that the mug is in the sink and the sink is empty (duplicate)
the goal is that the mug is in the sink and the sink is clean (duplicate)
\end{verbatim}

\begin{center}
\textbf{\emph{Repairing unknown terminology:}}
\end{center}

STARS continues by repairing another response, with a different mismatch, a response with an unknown word. In this response, shown below, the robot does not know the word "started".

\begin{verbatim}
the goal is that the mug is in the dishwasher and the dishwasher is
    started
\end{verbatim}

As before, the robot selects a template for repairing unknown terms, containing an example of an unknown term repair (shown below), and instantiates it with the relevant task context, the mismatched response, and the robot's repair response: ``No. Unknown word started.'' The prompt is shown below.

\begin{center}
\textbf{\emph{Prompt:}}
\end{center}
\begin{verbatim}
(EXAMPLES)(TASK)Task name: store object. Task context: I am in mailroom.
Aware of package of office supplies, package is in mailroom.
(RESULT)The goal is that the package is in the cabinet.(END RESULT)
Response: No. Unknown word cabinet.
(RESULT)The goal is that the package is in the closet.(END RESULT)
Response: Ok.
Steps:
1. Pick up package
2. Put package into closet
(END TASK)
(TASK)Task name: deliver package. Task context: I am in mailroom.
Aware of package addressed to Gary, package is in mailroom.
(RESULT)The goal is that the package is in Gary's office.(END RESULT)
Response: Ok.
Steps:
1. Pick up package addressed to Gary
2. Go to Gary's office
3. Put package onto desk in Gary's office
(END TASK)
(END EXAMPLES)
(TASK)Task name: tidy kitchen. Task context: I am in kitchen.
Aware of mug in dish rack.
(RESULT)the goal is that the mug is in the dishwasher and the dishwasher
is started(END RESULT)
Response: No. Unknown word started.
(RESULT)
\end{verbatim}

As before, this prompt is used to generate a set of responses using the ST beam search, producing the goal descriptions listed below. Note that the repaired responses no longer contain ``the dishwasher is started'' and contains other terms to describe the state of the dishwasher.

\begin{center}
\textbf{\emph{Final output for repair of unknown word started:}}
\end{center}
\begin{verbatim}
the goal is that the mug is in the dishwasher and the dishwasher is turned
    on (duplicate)
the goal is that the mug is in the dishwasher and the dishwasher is running
    (duplicate)
the goal is that the mug is in the dishwasher and the dishwasher is on
    (duplicate)
\end{verbatim}

In this case all these results are duplicates of ones found previously. 

\begin{center}
\textbf{\emph{Repairing an affordance mismatch:}}
\end{center}

Next the robot performs a repair for a response (shown below) with an affordance mismatch. In this case, the dish rack is not grabbable and therefore cannot be put into the cupboard.

\begin{verbatim}
the goal is that the mug is in the dish rack and the dish rack is in the
  cupboard
\end{verbatim}

The same process as before repeats, STARS selects a prompt template with an affordance repair example (shown below), instantiates with the task context, and provides the mismatched response and the robot's direction to repair the response: ``No. Rack is not grabbable.'' This prompt can be seen below.

\pagebreak
\begin{center}
\textbf{\emph{Prompt:}} 
\end{center}
\begin{verbatim}
(EXAMPLES)(TASK)Task name: store object. Task context: I am in mailroom.
Aware of package of office supplies, package is in mailroom.
(RESULT)The goal is that the package is on the shelf and the shelf is
on the table.(END RESULT)
Response: No. Shelf is not grabbable.
(RESULT)The goal is that the package is on the shelf.(END RESULT)
Response: Ok.
Steps:
1. Pick up package
2. Put package onto shelf
(END TASK)
(TASK)Task name: deliver package. Task context: I am in mailroom.
Aware of package addressed to Gary, package is in mailroom.
(RESULT)The goal is that the package is in Gary's office.(END RESULT)
Response: Ok.
Steps:
1. Pick up package addressed to Gary
2. Go to Gary's office
3. Put package onto desk in Gary's office
(END TASK)
(END EXAMPLES)
(TASK)Task name: tidy kitchen. Task context: I am in kitchen.
Aware of mug in dish rack.
(RESULT)the goal is that the mug is in the dish rack and the dish rack
is in the cupboard(END RESULT)
Response: No. Rack is not grabbable.
(RESULT)
\end{verbatim}

Performing tree retrieval using this prompt results in a pair of goal descriptions that do not have the affordance mismatch.

\begin{center}
\textbf{\emph{Final output for repair of affordance mismatch:}}
\end{center}
\begin{verbatim}
the goal is that the mug is in the dish rack
the goal is that the mug is in the cupboard (Duplicate)
\end{verbatim}

The responses generated through repair will be analyzed again by the robot to determine if they are viable, or if they contain mismatches. The robot will attempt to repair mismatched responses generated from a repair again. It will not attempt to repair a response for a third time; there needs to be some limit to prevent the robot from making continual repair prompts. STARS detects duplicates before sending them for analysis to the robot so multiple repairs will not be attempted on duplicate responses.

\subsection{Selection}

After performing Search Tree, Analysis, and Repair, the robot has generated a set of viable response for goal descriptions for the task to tidy a mug in the dish rack. These responses, ordered by mean log probability, are listed below:

\begin{center}
\textbf{\emph{Viable goal responses for a mug in the dish rack ordered by mean log probability}}
\end{center}
\begin{verbatim}
the goal is that the mug is in the cupboard (0.8612)
the goal is that the mug is in the dishwasher (0.8618)
the goal is that the mug is in the dishwasher and the dishwasher is closed
    (0.899)
the goal is that the mug is in the drawer and the drawer is closed
    (0.913)
the goal is that the mug is in the cupboard and the cupboard is closed
    (0.935)
the goal is that the mug is in the dish rack (0.971)
\end{verbatim}

Now the robot uses the LLM (in this case GPT-4) to select responses from the viable options by constructing a new prompt. It uses the selection prompt template and instantiates it with the candidate options and relevant task context.

The prompt and the response from GPT-4 using the LLM selection strategy are shown below. A small example prompt of this selection is presented in the beginning of the prompt (one-shot prompting). The prompt solicits a single token response, after ``Answer: '', from the LLM for an integer indicating which of the responses is the best. The options are presented in order by their mean log probability (lowest to highest). (GPT-4 appears to have a small bias toward selecting the most recently presented option, hence this ordering which biases towards the higher probability responses).  The order of options varies slightly between runs due to differences in the mean log probability calculated by the LLM. However, even with the temperature set to 0 and the same ordering of the same set of goals, there is occasional variance in the response.

\begin{center}
\textbf{\emph{Example selection prompt, including a prompt example}}
\end{center}
\begin{verbatim}
Task name: store object. Task context: I am in mailroom.
Aware of package on table.
Question: Which is the most reasonable goal for package on table?
1. The goal is that the package is on the floor.
2. The goal is that the package is in the closet.
Answer: 2.
Task name: tidy kitchen. Task context: I am in kitchen.
Aware of mug in dish rack.
Question: Which is the most reasonable goal for mug in dish rack?
1. The goal is that the mug is in the cupboard.
2. The goal is that the mug is in the dishwasher.
3. The goal is that the mug is in the dishwasher and the dishwasher is
    closed.
4. The goal is that the mug is in the drawer and the drawer is closed.
5. The goal is that the mug is in the cupboard and the cupboard is closed.
6. The goal is that the mug is in the dish rack.
Answer:
\end{verbatim}

\begin{center}
\textbf{\emph{The response from GPT-4 (Temperature=0):}}
\end{center}
\begin{verbatim}
5
\end{verbatim}

The response from the prompt for LLM selection chooses ``The goal is that the mug is in the cupboard and the cupboard is closed" as the best response for the goal for the mug in the dish rack. Without oversight the robot would select this goal description to learn from. In this case this is the correct goal, and shows benefit over the base line template-based prompting strategy of using the mean log probability, which would have selected an incorrect response of ``The goal is that the mug is in the dish rack.''

\subsection{Oversight}
To achieve the requirement of learning situational relevant knowledge, we need to be sure that the goal for each specific object conforms to the preferences of the human user.\footnote{The design of the system assumes individual users will have different preferences; i.e., one user may prefer that cereal is stored in the pantry and another may want it to be placed on the counter. However, the experimental design assumes a single ``user" with consistent preferences to make straightforward the assessment  of whether or not the simulated robot achieved this ``user's" desired goal state for each object.} STARS has produced a list of candidate goals, and used the LLM to select a preferred candidate. Neither the LLM nor the robot has knowledge of the preferences of this particular user in this particular selection, so confirmation by the user is required. To achieve this, the goal selected by STARS is now offered to the human for confirmation using this dialog:

\begin{quote}
\textbf{Robot:} [LM] For a mug in the dish rack is the goal is that the mug is in the cupboard and the cupboard is closed? \\
\textbf{Instructor}: yes.
\end{quote}

In this case the human responded in the affirmative. If the human responded negatively, the LLM Selection process would repeat, but with option 5 removed. This process repeats until the human confirms a goal as correct, the options produced from the LLM are exhausted, or the human is asked to confirm 5 different goal responses. Once these are exhausted, or the limit of questions is reached, the human is asked to describe the goal. This strategy of only asking the human for yes/no confirmations instead of asking for complete goal descriptions substantially reduces the amount of words required from the human to get 100\% task completion, as shown by our experimental results.

\section{Additional data analysis from experiments}
In this section, we present and describe in more detail the experimental results outlined in the main body of the paper. All experiments were run an virtual machine running on an HP laptop with an Intel Core i7 1165G7. The virtual machine, running Ubuntu 18.04, had access to 16 GB of ram and 4 cores.

Tables ~\ref{tab:app:extended-summary-tidy},~\ref{tab:app:extended-summary-groceries}, and ~\ref{tab:app:extended-summary-office}  present an extended summary of the data presented in the main body of the paper for the three tasks. The columns of the table are:

\begin{table}[th]
\centering
\begin{tabular}{rrrrrrrrrrr}
\hline
\rot{Condition} & \rot{Task Completion Rate ($\%$)} & \rot{Retrieved goals} & \rot{Proposed goals} & \rot{Sourced goals} & \rot{Total prompt tokens} & \rot{Total completion tokens} & \rot{Total tokens} & \rot{Total Instructions} & \rot{Total Yes/No Instructions} & \rot{Total user words} \\
\hline
TBP & 52.5 & 93 & 0 & 25 & 35,622 & 5,785 & 41,407 & 14 & 0 & 76 \\
TBP+O & 100.0 & 89 & 64 & 21 & 36,606 & 5,863 & 42,469 & 92 & 64 & 403 \\
ST & 50.0 & 243 & 0 & 24 & 55,491 & 1,383 & 56,874 & 14 & 0 & 76 \\
STS & 40.0 & 247 & 0 & 18 & 65,016 & 1,442 & 66,458 & 14 & 0 & 76 \\
STAR & 77.5 & 353 & 0 & 33 & 122,531 & 3,555 & 126,086 & 14 & 0 & 76 \\
STARS & 77.5 & 368 & 0 & 35 & 136,043 & 3,828 & 139,871 & 14 & 0 & 76 \\
STARS+O & 100.0 & 361 & 51 & 35 & 134,372 & 3,724 & 138,096 & 65 & 51 & 127 \\
\hline
\end{tabular}
    \caption{Extended summary of measures/condition for the ``tidy kitchen'' experiments.}
    \label{tab:app:extended-summary-tidy}
\end{table}

\begin{itemize}
    \item Condition: The experimental condition.
    \item Task Completion Rate: The fraction of the task completed by the agent in the condition. In the ``tidy kitchen'' experiment, there are 35 objects with a desired final location (see Table~\ref{tab:objects}) and 5 kitchen locations with a desired final state (such as ``refrigerator door closed"; see Table~\ref{tab:kitchen-objects}). Task completion rate is computed as the fraction of these 40 assertions that match the desired final state. For the ``store groceries'' experiment, there are 15 objects with a desired final location and 3 objects with a desired final state of closed. For the ``organize office'' task, there are 12 objects with a desired final location and 2 objects with a desired final state of closed.
    \item Retrieved goals: The total number of goals generated by the LLM. A retrieved goal is produced an invocation of Template-based Prompting (baseline conditions) or Search Tree (STARS conditions, including use of Search Tree in Analysis and Repair).
    \item Proposed Goals: The total number of goals presented (``proposed" as an option) to the user in the oversight conditions.
    \item Sourced Goals: The number of proposed goals that are actually used (or ``sourced") by the robot. When the agent can recognize that a goal is unviable, it does not attempt to use that goal, which explains why some non-oversight conditions have less than 35 (tidy), 15 (store), or 12 (organize) goals respectively. In addition, for TBP+0, for ``tidy kitchen'' only 21 goals could be sourced (meaning that the user had to provide descriptions for 14 of the objects in the kitchen). For TBP+0, for ``store groceries'' 13 goals could be sourced, and for ``organize office'' only 5 goals could be sourced.
    \item Total prompt tokens: The total number of tokens sent to a LLM for the condition. Total tokens includes tokens sent for both Search Tree (including ST under AR) and Selection. 
    \item Total completion tokens: The total number of tokens received from the LLM for the condition.
    \item Total tokens: The sum of total prompt tokens and completion tokens.
    \item Total instructions: The total number of instructions provided to the robot for that condition. In the non-oversight (as well as oversight conditions), the user provides some initial instructions (e.g., tidy kitchen by clearing the table, etc.) as well as confirmation of the completion of tasks, resulting in a floor of 14 instructions (tidy kitchen), 6 instructions (store groceries, organize office). On the oversight conditions, total instructions includes any goal descriptions that the user provides (``the goal is that the steak knife is in the dishwasher") as well as confirming/disconfirming feedback (Agent: ``Is that goal that the steak knife is in the cupboard?". User: ``No.") 
    \item Total Yes/No Instructions: The number of yes/no feedback responses provided by the user in the oversight conditions.
    \item Total user words: The total number of user words provided to the robot for that condition during the experiment. Using the examples under ``Total Instructions," the goal description is 11 words and the yes/no question would be a single word for those instructions.
\end{itemize}

\begin{table}[th]
\centering
\begin{tabular}{rrrrrrrrrrr}
\hline
\rot{Condition} & \rot{Task Completion Rate ($\%$)} & \rot{Retrieved goals} & \rot{Proposed goals} & \rot{Sourced goals} & \rot{Total prompt tokens} & \rot{Total completion tokens} & \rot{Total tokens} & \rot{Total Instructions} & \rot{Total Yes/No Instructions} & \rot{Total user words} \\
\hline
TBP & 66.7 & 39 & 0 & 13 & 14,878 & 2,120 & 17,078 & 6 & 0 & 28 \\
TBP+O & 1.0 & 37 & 21 & 13 & 16,362 & 2,327 & 18,689 & 29 & 21 & 92 \\
ST & 66.7 & 96 & 0 & 13 & 20,932 & 586 & 21,518 & 6 & 0 & 28 \\
STS & 66.7 & 99 & 0 & 9 & 25,085 & 605 & 25,690 & 6 & 0 & 28 \\
STAR & 77.8 & 170 & 0 & 15 & 56,005 & 1,704 & 57,709 & 6 & 0 & 28 \\
STARS & 94.4 & 171 & 0 & 15 & 60,069 & 1,739 & 61,808 & 6 & 0 & 28 \\
STARS+O & 100.0 & 177 & 16 & 15 & 62,693 & 1,808 & 64,501 & 22 & 16 & 44 \\
\hline
\end{tabular}
    \caption{Extended summary of measures/condition for the ``store groceries'' experiments.}
    \label{tab:app:extended-summary-groceries}
\end{table}

\begin{table}[th]
\centering
\begin{tabular}{rrrrrrrrrrr}
\hline
\rot{Condition} & \rot{Task Completion Rate ($\%$)} & \rot{Retrieved goals} & \rot{Proposed goals} & \rot{Sourced goals} & \rot{Total prompt tokens} & \rot{Total completion tokens} & \rot{Total tokens} & \rot{Total Instructions} & \rot{Total Yes/No Instructions} & \rot{Total user words} \\
\hline
TBP & 35.7 & 34 & 0 & 5 & 11,232 & 1,690 & 12,992 & 6 & 0 & 28 \\
TBP+O & 1.0 & 35 & 28 & 5 & 9,996 & 1,666 & 11,662 & 41 & 28 & 184 \\
ST & 21.4 & 95 & 0 & 3 & 20,641 & 441 & 21,082 & 6 & 0 & 28 \\
STS & 21.4 & 97 & 0 & 1 & 24,256 & 461 & 24,717 & 6 & 0 & 28 \\
STAR & 64.3 & 204 & 0 & 12 & 73,357 & 2,152 & 75,509 & 6 & 0 & 28 \\
STARS & 92.9 & 201 & 0 & 12 & 73,933 & 2,123 & 76,056 & 6 & 0 & 28 \\
STARS+O & 100.0 & 206 & 15 & 11 & 75,554 & 2,168 & 77,722 & 22 & 15 & 60 \\
\hline
\end{tabular}
    \caption{Extended summary of measures/condition for the ``organize office'' experiments.}
    \label{tab:app:extended-summary-office}
\end{table}

\begin{figure}[thb]
    \centering
\includegraphics[width=.6\linewidth]{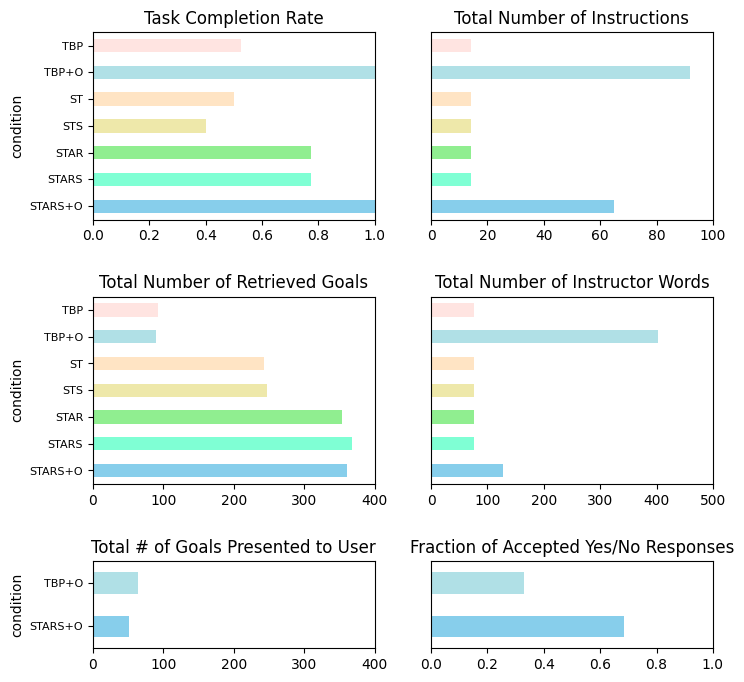}
    \caption{Expanded panel of summary results from the ``tidy kitchen'' experiment.}
    \label{fig:app:summary-matrix}
\end{figure}

Figure~\ref{fig:app:summary-matrix} presents an expanded summary of key results for the ``tidy kitchen'' task from Figure~\ref{fig:overall_results_typical} in the main body of the paper. %
Task completion, total number of instructor words, and fraction of accepted yes/no responses are discussed in the main body of the paper. 
\begin{itemize}
    \item Total number of instructions: Similar to total number of instructor words, total number of instructions decreases in the STARS oversight condition in comparison to template-based prompting. 65 interactions are needed. However, 51 of these interactions are proposed goals that require yes/no responses and 35 of these are accepted (68\% acceptance rate, as in the lower right chart). Note that in the STARS+O condition, there was at least one acceptable goal condition generated by the LLM for each object in the data set.
    \item Number of Retrieved Goals: This chart compares how many goal descriptions are retrieved from the LLM. In the TBP conditions, relatively few goal descriptions are produced ($\sim$90, or about 2.6 descriptions/object). With the ST conditions, many more goals are retrieved ($\sim$245) due to beam search. In the STAR+ conditions, about 365 goals are retrieved. The increase of about 120 goal retrievals represents the additional LLM retrievals being performed by beam search as part of Analysis and Repair.
    \item Total Goals Presented to User: This chart illustrates the number of retrieved goals presented to the user (both charts share the same horizontal axis). In the TBP+O condition, 64 of the 89 retrieved goals are presented to the user (and only 21 are eventually used by the robot). In the STARS+O condition, slightly fewer goals are presented (51) from the total of 361 goals retrieved and one goal is used for each object (35 sourced goals). This result highlights the while the retrieval process is much broader for STARS than for TBD, the search and evaluation processes result in greater overall precision in identifying acceptable goal descriptions, requiring fewer user evaluations and producing a higher acceptance rate when a goal needs to be confirmed (oversight).
    \end{itemize}

\begin{figure}[htb]
    \centering
    \includegraphics[width=.6\linewidth]{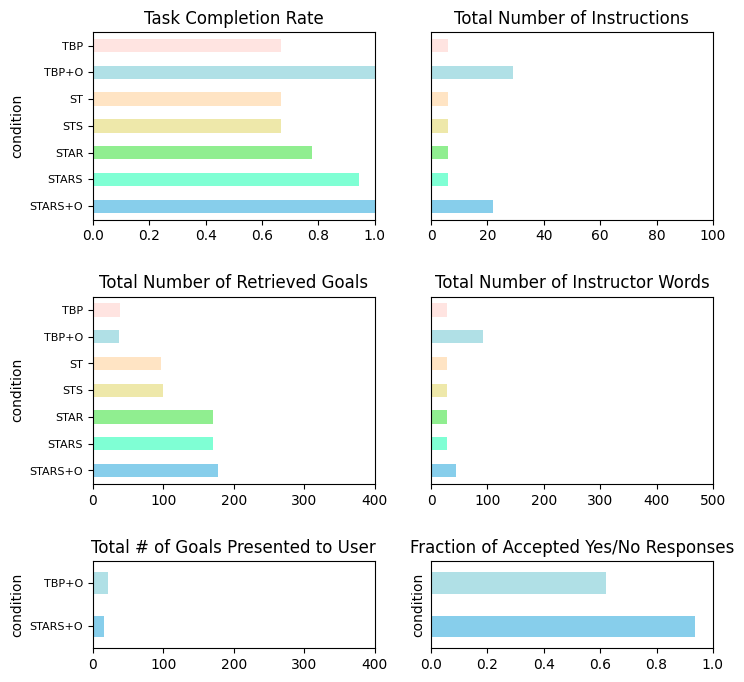}
    \caption{Performance and user cost measures for experimental conditions for ``store groceries'' task.}
    \label{fig:app:overall_results_groceries}
\end{figure}

Figure~\ref{fig:app:overall_results_groceries} presents a summary of key results for the ``store groceries'' task. Details for ``store groceries'' for measures not discussed in the main body of the paper are as follows.

\begin{itemize}
    \item Total number of instructions: Total number of instructions decreases in the STARS oversight condition in comparison to TBP. 22 interactions are needed, but 16 of these interactions are proposed goals that require yes/no responses and 15 of these are accepted (94\% acceptance rate, as in the lower right chart). In the STARS+O condition, at least one acceptable goal condition was generated by the LLM for each object in the data set.
    \item Number of Retrieved Goals: In the TBP conditions, few goal descriptions are produced (39, or 2.6 descriptions per object on average). With the ST conditions, many more goals are retrieved (96). In the STAR+ conditions, 170-177 goals are retrieved. The increase of $\sim$80 goal retrievals is due to additional LLM retrievals from beam search using during repairs of Analysis and Repair.
    \item Total Goals Presented to User: In the TBP+O condition, 21 of the 37 retrieved goals are presented to the user (and only 13 are used by the robot). In the STARS+O condition, slightly fewer goals are presented (16) from the total of 177 goals retrieved and one goal is used for each object (15 sourced goals). This result highlights again that the Search Tree and Analysis processes result in greater overall precision in identifying acceptable goal descriptions, requiring fewer user evaluations and generating a higher acceptance rate when goals need to be confirmed (using oversight).
\end{itemize}

\begin{figure}[thb]
    \centering
    \includegraphics[width=.6\linewidth]{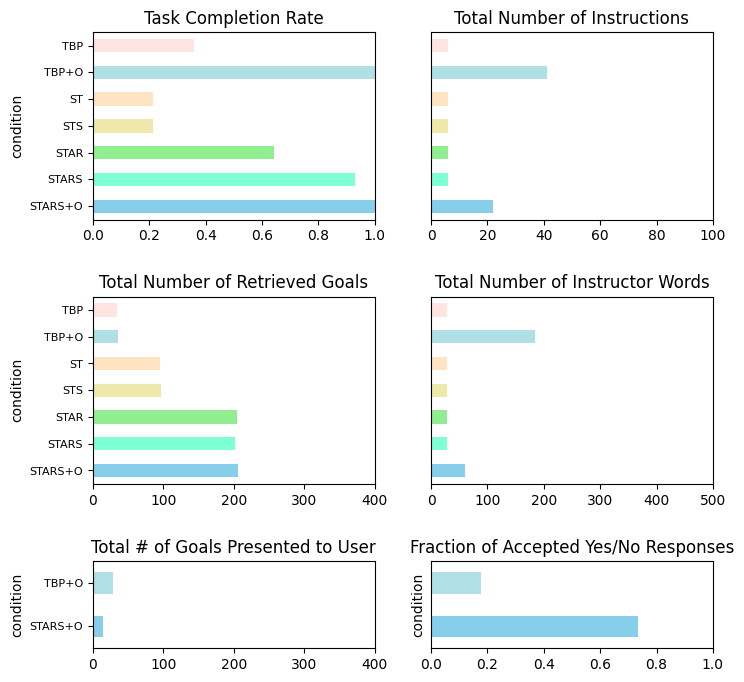}
    \caption{Performance and user cost measures for experimental conditions for the ``organize office'' task.}
    \label{fig:app:overall_results_office}
\end{figure}

Figure~\ref{fig:app:overall_results_office} presents a summary of key results for the ``organize office'' task. Details for ``organize office'' are as follows.

\begin{itemize}
    \item Total number of instructions: As with the other tasks, the total number of instructions decreases in the STARS oversight condition compared to TBP. With STARS 22 interactions are needed, but 15 of these interactions are goal proposals that require yes/no responses and 11 of these are accepted (73\% acceptance rate, as in the lower right chart).
    \item Number of Retrieved Goals: In the TBP conditions, as shown in other tasks, relatively few goal descriptions are produced (34, or 2.8 descriptions per object). With the ST conditions, many more goals are retrieved (95) from the beam search. In the STAR+ conditions, $\sim$205 goals are retrieved. Again, the increase of goal retrievals ($\sim$110) is due to the additional LLM retrievals being performed by beam search as part of Analysis and Repair.
    \item Total Goals Presented to User: In the TBP+O condition, 28 of the 35 retrieved goals are presented to the user, but only 5 are used by the robot. In the STARS+O condition, fewer goals are presented (15) from the total of 206 goals retrieved and almost one goal is used for each object (11 sourced goals). The user had to be queried for a goal for one of the objects. As showed with the other tasks, the retrieval process is much broader for STARS than for TBP, but the ST and AR processes result in greater overall precision in identifying acceptable goal descriptions, requiring fewer user evaluations and creating a higher acceptance rate with oversight.
\end{itemize}

\begin{figure}
\centering
\begin{subfigure}{.33\textwidth}
  \centering
  \includegraphics[width=1.0\linewidth]{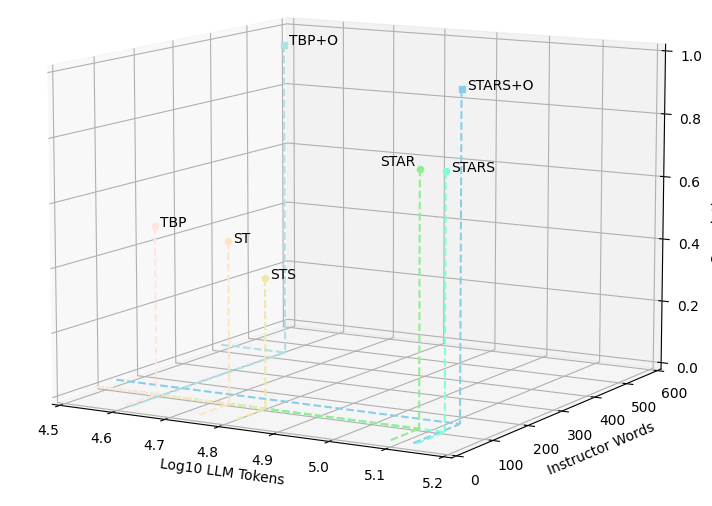}
  \caption{Tidy kitchen.}
  \label{fig:token_words_completion-tidy}
\end{subfigure}%
\begin{subfigure}{.33\textwidth}
  \centering
  \includegraphics[width=1.0\linewidth]{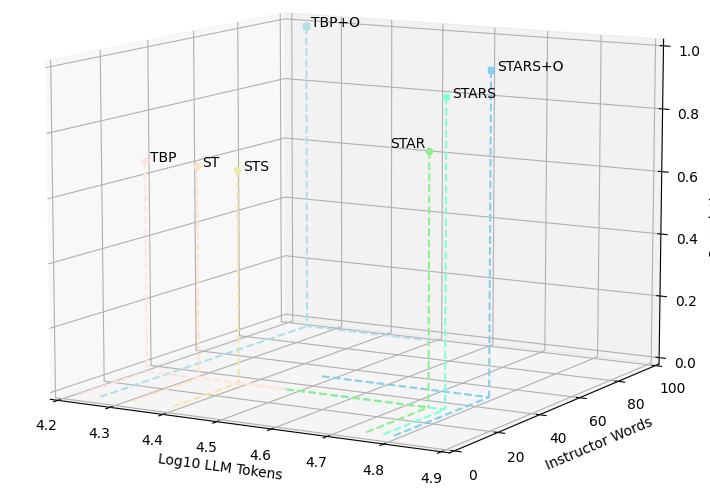}
  \caption{Store groceries.}
  \label{fig:token_words_completion-groceries}
\end{subfigure}
\begin{subfigure}{.33\textwidth}
  \centering
  \includegraphics[width=1.0\linewidth]{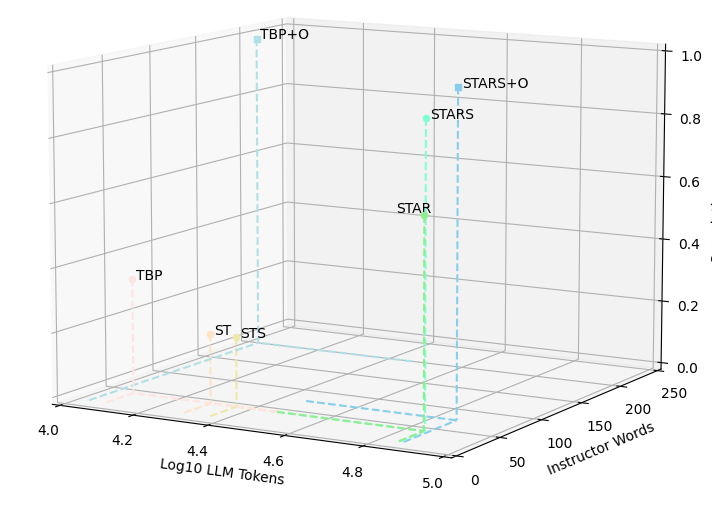}
  \caption{Organize office.}
  \label{fig:token_words_completion-office}
\end{subfigure}
\caption{Number of log$_{10}$ tokens vs. words vs. task completion rate for all experimental conditions for the three tasks.}
\label{fig:token_words_completion_all}
\end{figure}

Figure \ref{fig:token_words_completion_all} shows the trade off between the costs (words and tokens) and performance (task completion) and highlights the relative contributions of the components of the STARS strategy for the three tasks. Figure~\ref{fig:token_words_completion-tidy} show the trade off for the ``tidy kitchen'' task. For this tasks, Search Tree (ST) and Analysis and Repair (AR) have the largest impact on token cost. The benefits in performance are not observed until adding Analysis and Repair that down-selects from the now larger space of responses. The figure also shows that STARS greatly reduces the human cost in words (while increasing token costs), and shows that Selection doesn't have an appreciable impact on performance for this task.

Figure \ref{fig:token_words_completion-groceries} shows the cost/performance trade off for the ``store groceries'' task. For this task, Search Tree has a smaller impact on token cost. Adding Analysis and Repair (AR) has a larger impact on token cost, but as before, increases performance significantly. The figure shows again that STARS greatly reduces the human cost in words (while increasing token costs), but in this case Selection does have an appreciable impact on performance.

Figure \ref{fig:token_words_completion-office} shows the  the cost/performance trade off for the ``organize office task'' task. For this task, Search Tree has a compartively larger impact on token cost, while Adding Analysis and Repair (AR) has a much larger impact. As shown in the other tasks, AR increases performance by a large amount. The figure shows again that STARS greatly reduces the human cost in words, and as with the ``store groceries'' tasks, Selection has an large impact on performance, showing an increase from 64\% (STAR) to 93\% (STARS).

\begin{figure}[th]
    \centering
    \includegraphics[width=.6\columnwidth]{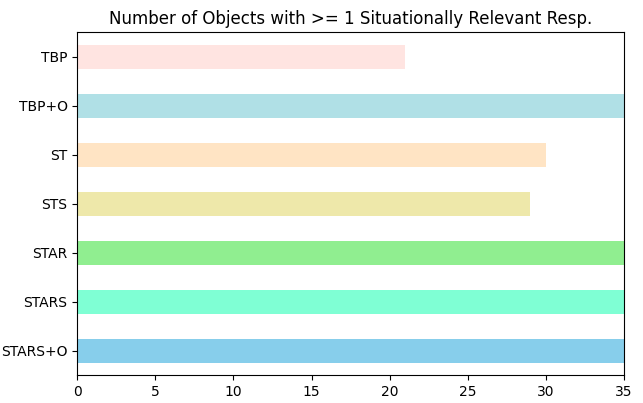}
    \caption{Evaluating performance of STARS in terms of individual objects (left).}
    \label{fig:objects_situational_rel}
\end{figure}

Figure \ref{fig:objects_situational_rel} shows for each condition for the ``tidy kitchen'' task, the number of objects (out of 35) for which the robot retrieved at least one situationally relevant response from the LLM. While only retrieving situationally responses for 15 objects in the baseline, STARS results in 100\% of the objects having situationally relevant responses, largely due to the Search Tree and Analysis and Repair. This chart illustrates that the STARS strategy is successful at generating situationally relevant responses from the robot, even if those responses are not always selected first by the robot. 

\begin{figure}[ht]
    \centering
    \includegraphics[width=0.45\columnwidth]{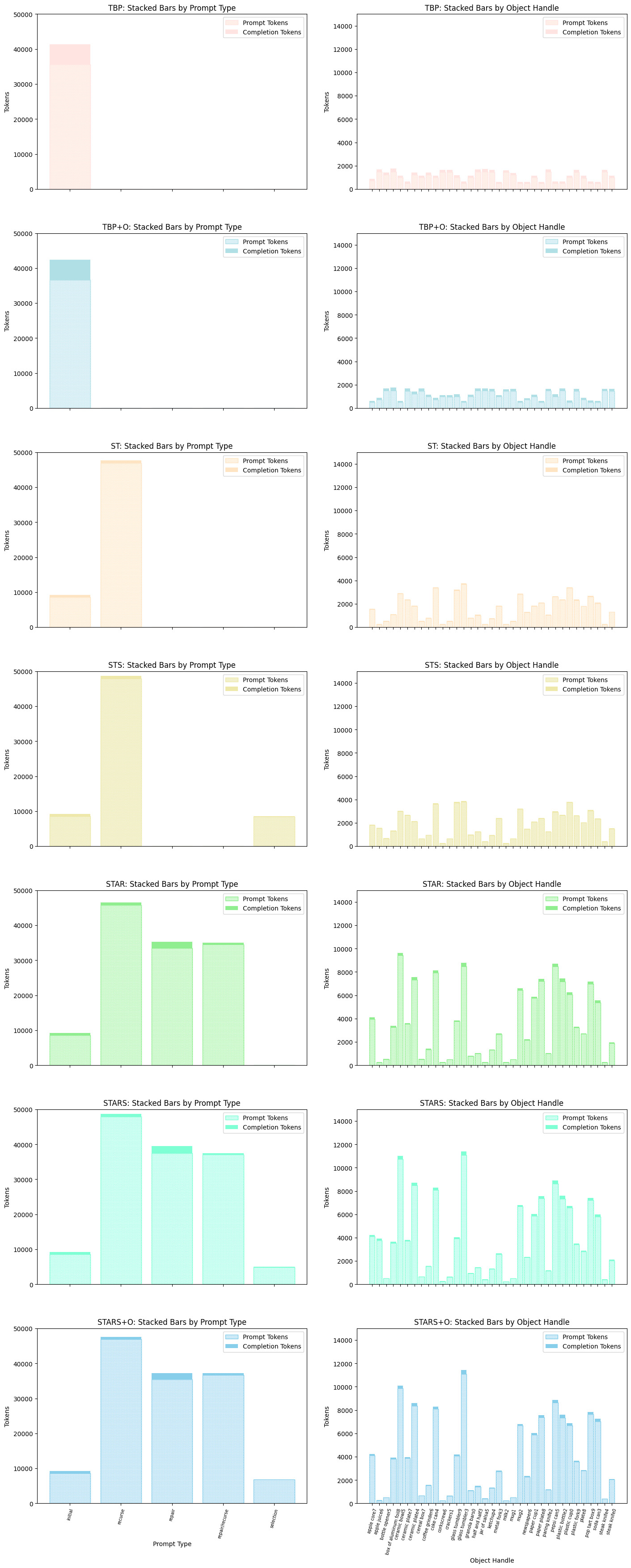}
    \caption{Detailed summary of token usage by prompt type (left) and for individual objects (right) for the ``tidy kitchen'' task. The hatched areas summarize the prompts sent to the LLM and the solid areas the number of tokens received in response to those prompts.}
    \label{fig:detailed-token-analysis}
\end{figure}

Figure~\ref{fig:detailed-token-analysis} shows the token cost (from prompts and generation) for each experimental condition for the ``tidy kitchen'' task, showing the tokens used per object (left) and the tokens used for each prompt type. Some objects, particularly in the conditions with analyze and repair, result in many more tokens being used. The types of prompts (in order left to right) include the initial prompt, recursive (prompts used for the Search Tree beam search), repair (prompts using during Analysis and Repair), repair/recurse (prompt used for beam search during repair), and selection (prompt used for LLM Selection over candidates). Based on the condition, only certain types of prompts are used.

\begin{figure}[ht]
    \centering
    \includegraphics[width=0.45\columnwidth]{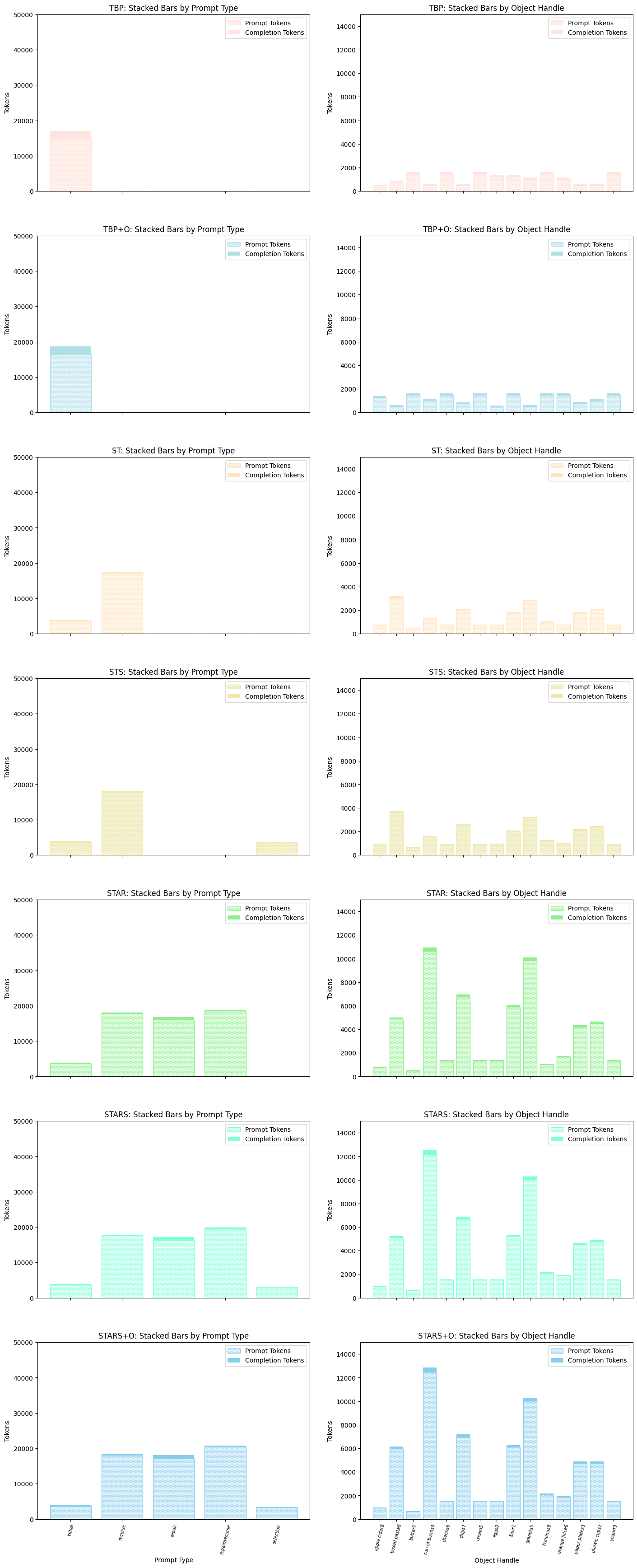}
    \caption{Detailed summary of token usage by prompt type (left) and for individual objects (right) for the ``store groceries'' task.. The hatched areas summarize the prompts sent to the LLM and the solid areas the number of tokens received in response to those prompts.}
    \label{fig:detailed-token-analysis-groceries}
\end{figure}

\begin{figure}[ht]
    \centering
    \includegraphics[width=0.45\columnwidth]{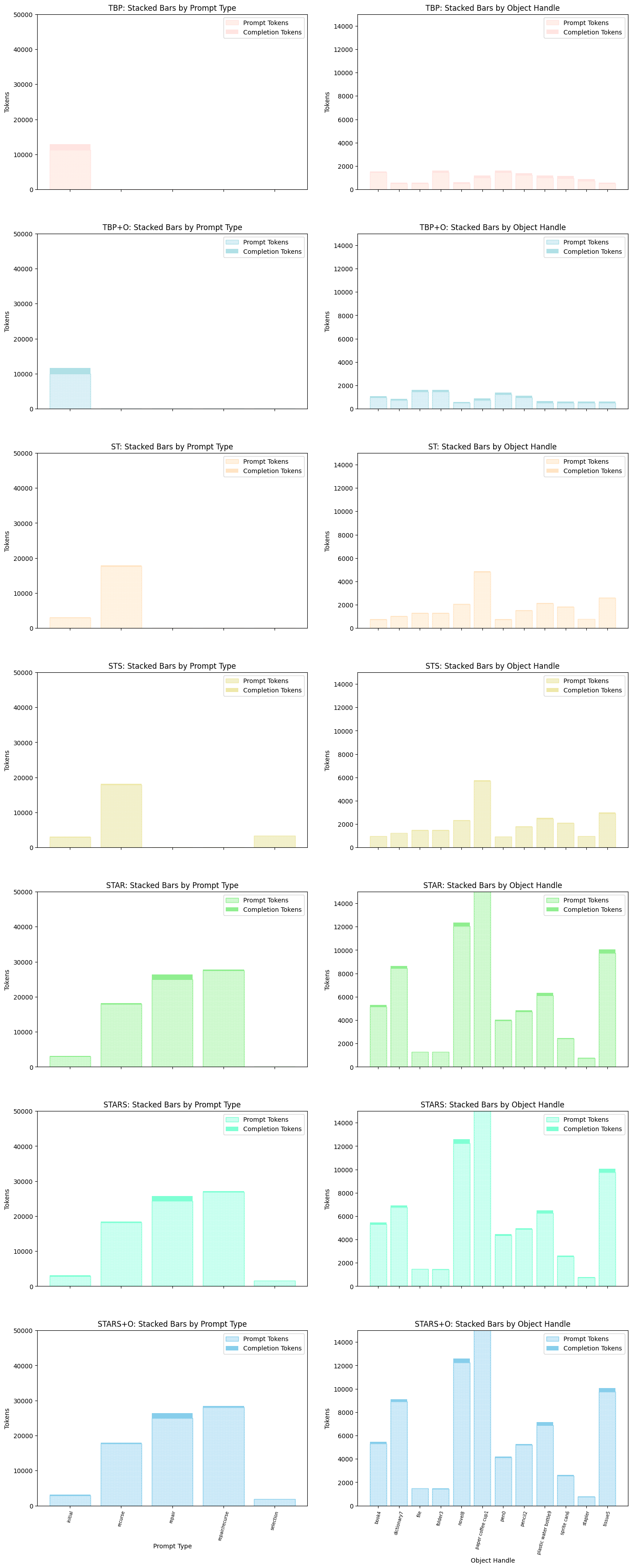}
    \caption{Detailed summary of token usage by prompt type (left) and for individual objects (right) for the ``organize office'' task.. The hatched areas summarize the prompts sent to the LLM and the solid areas the number of tokens received in response to those prompts.}
    \label{fig:detailed-token-analysis-office}
\end{figure}

Figures~\ref{fig:detailed-token-analysis-groceries} and ~\ref{fig:detailed-token-analysis-office} shows the token cost (from prompts and generation) for each experimental condition for the ``store groceries'' task. The results for these tasks are consistent with the ``tidy kitchen'' task.

\begin{figure}[ht]
    \centering
    \includegraphics[width=0.65\columnwidth]{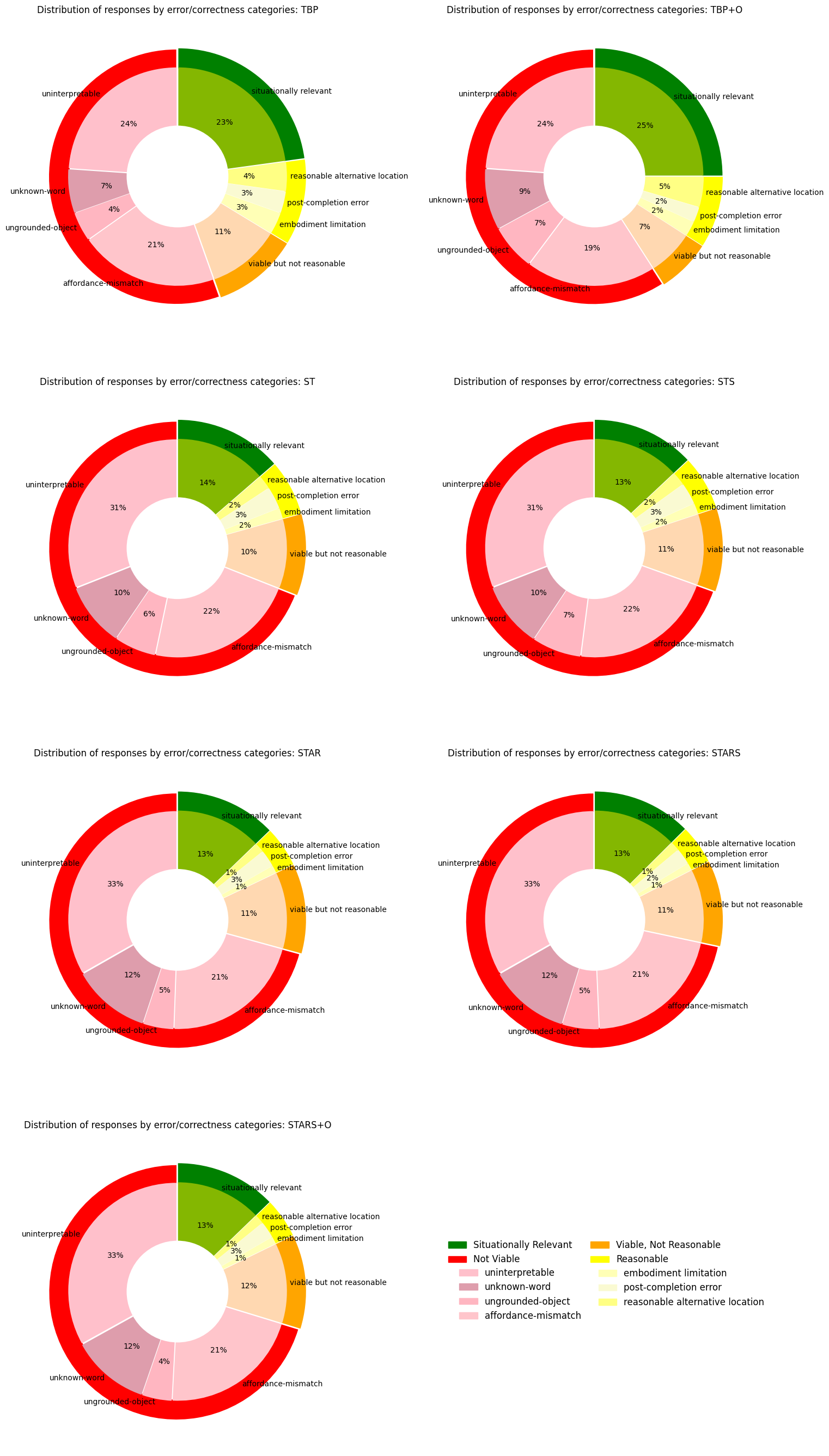}
    \caption{Categorization of all LLM responses for the experimental conditions for ``tidy kitchen'' task. These charts illustrate the distribution of various categories of responses over all the LLM responses produced. Primary categories are: not viable, viable but not reasonable, reasonable but not situationally relevant, and situationally relevant. Further sub-categorization of responses is shown for the not viable and reasonable categories.}
    \label{fig:detailed-response-categories}
\end{figure}

Figure~\ref{fig:detailed-response-categories} shows the categorization of LLM responses according to viability, reasonableness, and situational relevance for every experimental condition for the ``tidy kitchen'' task. As outlined in the paper, the distribution of responses in the ST-AR-S conditions are quite similar, in contrast to the baseline conditions (TBP and TBP+O) which reveal a different pattern. The baseline conditions show more situationally relevant responses by percentage, but many fewer responses are retrieved in these conditions. STARS results in an increase in the total number of situationally relevant responses retrieved, at the cost of generating more unviable responses (by percentage) overall. 

\begin{figure}[ht]
    \centering
    \includegraphics[width=0.65\columnwidth]{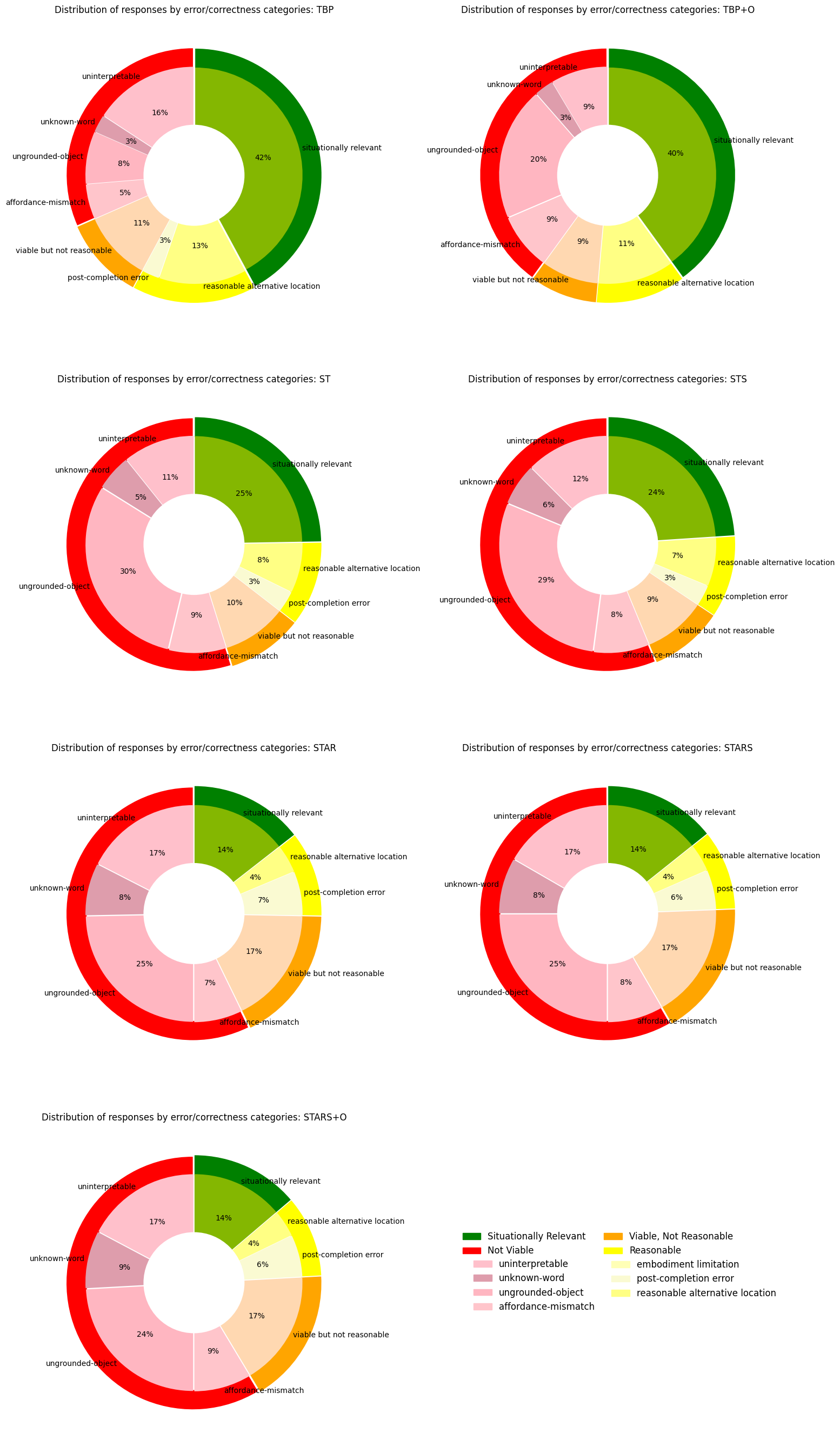}
    \caption{Categorization of all LLM responses for the experimental conditions for ``store groceries'' task. These charts illustrate the distribution of various categories of responses over all the LLM responses produced. Primary categories are: not viable, viable but not reasonable, reasonable but not situationally relevant, and situationally relevant. Further sub-categorization of responses is shown for the not viable and reasonable categories.}
    \label{fig:detailed-response-categories-groceries}
\end{figure}

Figure~\ref{fig:detailed-response-categories-groceries} shows the categorization of LLM responses according to viability, reasonableness, and situational relevance for every experimental condition for the ``store groceries'' task. The distributions of responses are similar to that from the ``tidy kitchen'' tasks, but with an increase across conditions of the percentage of situationally relevant responses and a decrease across conditions in the percentage of not viable responses. This is likely due to the task being simpler than ``tidy kitchen.''

\begin{figure}[ht]
    \centering
    \includegraphics[width=0.638\columnwidth]{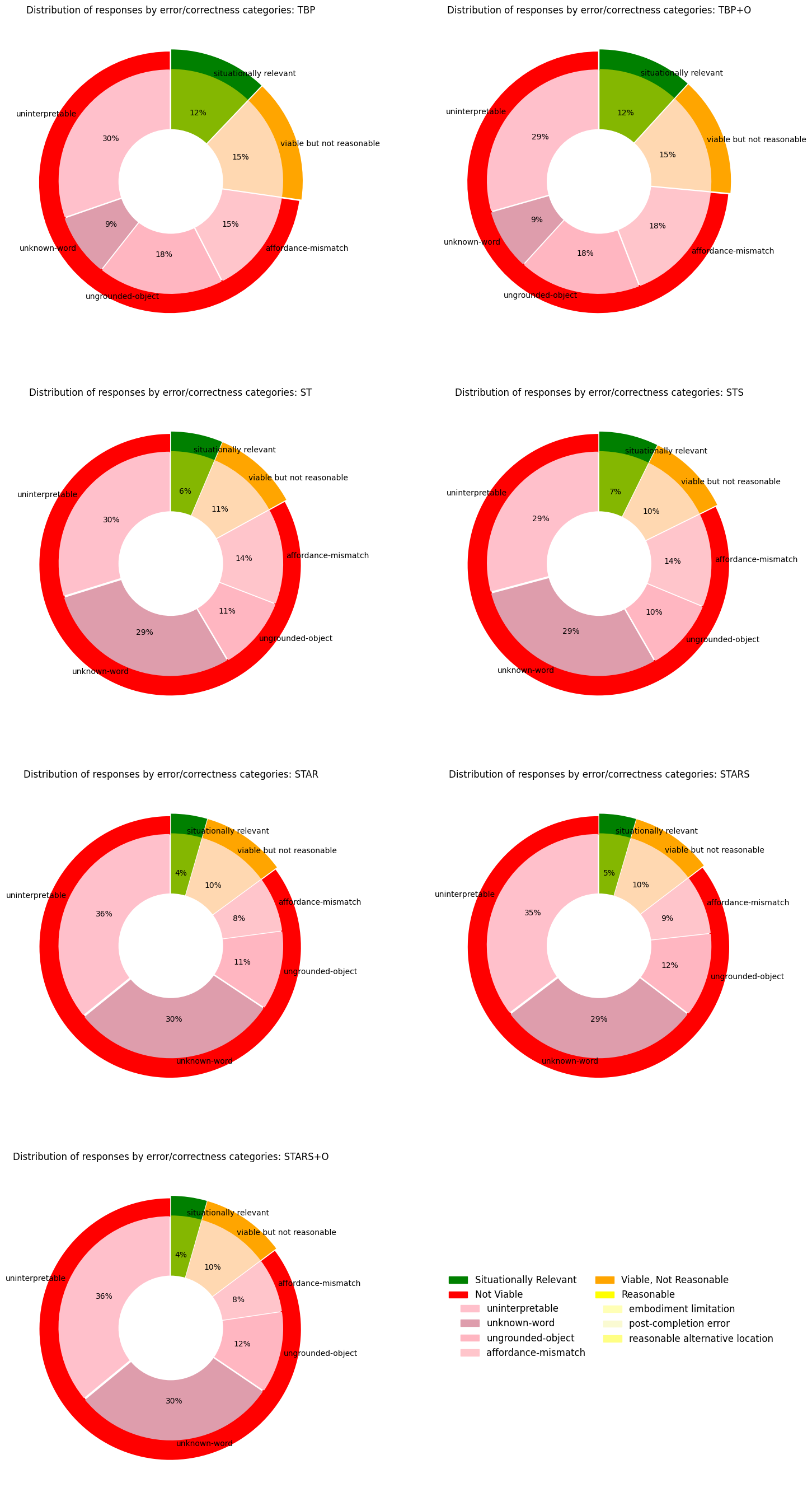}
    \caption{Categorization of all LLM responses for the experimental conditions for ``organize office'' task. These charts illustrate the distribution of various categories of responses over all the LLM responses produced. Primary categories are: not viable, viable but not reasonable, reasonable but not situationally relevant, and situationally relevant. Further sub-categorization of responses is shown for the not viable and reasonable categories.}
    \label{fig:detailed-response-categories-office}
\end{figure}

Figure~\ref{fig:detailed-response-categories-office} shows the categorization of LLM responses according to viability, reasonableness, and situational relevance for every experimental condition for the ``organize office'' task. The distributions of responses, compared to the prior two tasks, show a decrease across conditions of the percentage of situationally relevant responses and an increase across conditions in the percentage of not viable responses. From inspection of responses, this was due to many responses not being aligned with the specific office that the agent was situated in (e.g., referring to desk drawers instead of drawers).

\clearpage
\section{Exploration of Variability}

As mentioned in the body of the paper, there is little variation from one run to another of the same condition  (although there is slightly more variation in the 
\texttt{tidy kitchen} task in comparison to the other two tasks). This section of the appendix further explores what variability there is. Because running the experiment is somewhat expensive in time (especially in the oversight conditions) and not trivially inexpensive in the financial costs of LLM use, given the limited variability of the consequent results, we ran all conditions for the primary experiment only once.

\begin{table}[htbp]
\centering
\begin{tabular}{rrrrrrrrrrr}
\hline
\rot{Condition} & \rot{Task Completion Rate ($\%$)} & \rot{Retrieved goals} & \rot{Proposed goals} & \rot{Sourced goals} & \rot{Total prompt tokens} & \rot{Total completion tokens} & \rot{Total tokens} & \rot{Total Instructions} & \rot{Total Yes/No Instructions} & \rot{Total user words} \\
\hline
\multicolumn{11}{c}{\texttt{tidy kitchen}} \\
\hline
run1 & 77.5 & 360 & -- & 35 & 130,950 & 3,682 & 134,632 & 14 & -- & 76 \\
run2 & 75.0 & 347 & -- & 35 & 125,666 & 3,552 & 129,218 & 14 & -- & 76 \\
run3 & 77.5 & 357 & -- & 35 & 128,841 & 3,605 & 132,446 & 14 & -- & 76 \\
run4 & 75.0 & 355 &-- & 35 & 130,476 & 3,674 & 134,150 & 14 & -- & 76 \\
run5 & 75.0 & 354 & -- & 35 & 128,255 & 3,633 & 131,888 & 14 & -- & 76 \\
run6 & 80.0 & 364 & -- & 35 & 133,645 & 3,728 & 137,373 & 14 & -- & 76 \\
run7 & 80.0 & 359 & -- & 35 & 130,657 & 3,666 & 134,323 & 14 & -- & 76 \\
run8 & 77.5 & 357 & -- & 35 & 130,082 & 3,647 & 133,729 & 14 & -- & 76 \\
run9 & 80.0 & 353 & -- & 35 & 130,521 & 3,658 & 134,179 & 14 & -- & 76 \\
run10 & 77.5 & 355 & -- & 35 & 129,067 & 3,594 & 132,661 & 14 & -- & 76 \\
\hline
Mean & 77.5 & 356 & -- & -- & 129,816 & 3,643 & 133,459 & -- & -- & -- \\
Std. Dev. & 2.04 & 4.5 & -- & -- & 2,077 & 50 & 2,124 & -- & -- & -- \\
\hline
\hline
\multicolumn{11}{c}{\texttt{store groceries}} \\
\hline
run1 & 94.4 & 171 & -- & 15 & 60,069 & 1,739 & 61,808 & 6 & -- & 28 \\
run2 & 94.4 & 173 & -- & 15 & 60,443 & 1,683 & 62,126 & 6 & -- & 28 \\
run3 & 94.4 & 175 & -- & 15 & 60,784 & 1,675 & 62,459 & 6 & -- & 28 \\
run4 & 94.4 & 176 & -- & 15 & 60,558 & 1,720 & 62,278 & 6 & -- & 28 \\
run5 & 94.4 & 178 & -- & 15 & 60,990 & 1,710 & 62,700 & 6 & -- & 28 \\
run6 & 94.4 & 176 & -- & 15 & 61,041 & 1,697 & 62,738 & 6 & -- & 28 \\
run7 & 94.4 & 177 & -- & 15 & 61,321 & 1,706 & 63,027 & 6 & -- & 28 \\
run8 & 94.4 & 179 & -- & 15 & 61,620 & 1,707 & 63,327 & 6 & -- & 28 \\
run9 & 88.9 & 178 & -- & 15 & 62,502 & 1,730 & 64,232 & 6 & -- & 28 \\
run10 & 94.4 & 177 & -- & 15 & 62,222 & 1,737 & 63,959 & 6 & -- & 28 \\
\hline
Mean & 93.89 & 176 & -- & -- & 61,115 & 1,710 & 62,865 & -- & -- & -- \\
Std. Dev. & 1.76 & 2.4 & -- & -- & 776 & 21.6 & 783 & -- & -- & -- \\
\hline
\hline
\multicolumn{11}{c}{\texttt{organize office}} \\
\hline
run1 & 92.9 & 201 & -- &  12& 73,933 & 2,123 & 76,056 &  6& -- &  28\\
run2 & 92.9 & 200 & -- &  12& 73,355 & 2,128 & 75,483 &  6& -- &  28\\
run3 & 92.9 & 205 & -- &  12& 74,958 & 2,164 & 77,122 &  6& -- &  28\\
run4 & 92.9 & 200 & -- &  12& 73,020 & 2,126 & 75,146 &  6& -- &  28\\
run5 & 92.9 & 200 & -- &  12& 73,944 & 2,154 & 76,098 &  6& -- &  28\\
run6 & 92.9 & 205 & -- &  12& 75,134 & 2,159 & 77,293 &  6& -- &  28\\
run7 & 92.9 & 197 & -- &  12& 72,746 & 2,111 & 74,857 &  6& -- &  28\\
run8 & 92.9 & 207 & -- &  12& 75,852 & 2,182 & 78,034 &  6& -- &  28\\
run9 & 85.7 & 207 & -- &  12& 75,216 & 2,167 & 77,383 &  6& -- &  28\\
run10 & 92.9 & 204 & -- &  12& 75,212 & 2118 & 77,330 &  6& -- &  28\\
\hline
Mean & 92.14 & 202 & -- & -- & 74,337 & 2,143 & 76,480 & -- & -- & -- \\
Std. Dev. & 2.26 & 3.4 & -- & -- & 1,075 & 24.7 & 1093 & -- & -- & -- \\
\hline
\end{tabular}
    \caption{Measures for the STARS condition over ten runs for the three experimental tasks.}
    \label{tab:app:variability}
\end{table}

\begin{figure}[htbp]
    \centering
    \includegraphics[width=0.65\linewidth]{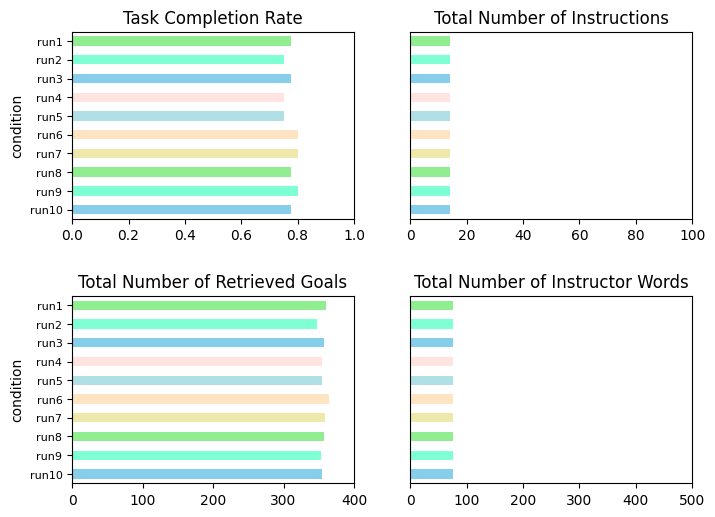}
    \caption{Comparing the variation of  outcomes over 10 STARS runs for the \texttt{tidy kitchen} task.}
    \label{fig:app:summary-matrix_variability}
\end{figure}

Table~\ref{tab:app:variability} shows the detailed summary of measures for 10 runs of the STARS condition (no oversight) for all three of the tasks. Two additional lines summarize with mean and standard deviation for those data that vary in the STARS condition. The table follows the format of Table~\ref{tab:app:extended-summary-tidy} and the definition of the individual measures are summarized in that table. Because STARS is not an oversight condition, the total number of instructions and total words do not change from run to run. Similarly, no goals are proposed to the user and thus there are no yes/no responses to those proposed goals. The results for \texttt{tidy kitchen} are also illustrated graphically in Figure~\ref{fig:app:summary-matrix_variability}.

As these results show, there is little change in overall results from run to run. In \texttt{tidy kitchen}, the Task Completion Rate varies from 75\% to 80\%, or from 30 to 32 of the 40 state assertions defined for the final desired state. There are even smaller variations (in a relative sense) in the retrieval and token measures. In all 10 conditions, STARS produces a viable goal that is sourced by the robot to execute.

While the lack of variability may appear unexpected, it is actually a consequence of the LLM's embedded token probabilities (which are fixed once the LLM is trained) and the experimental design, in which an object's gross location (``plate on the table" rather than a specific location on the table) is used for prompt generation. For any given object that the robot perceives, it will generate an instantiated prompt from the goal-description template using the gross location (``location: table").\footnote{In other work, we have explored the effects of the number of examples for few-shot, in-context learning with template-based prompting, as well as analysis of how well particular prompt examples contribute to the four main requirements. However, for this experiment, we used a single, fixed example in all prompt templates, which means that for a given object in a gross location, the prompt will be exactly the same for that object.}

While the task completion results from the other two tasks identical for all but one run, there is somewhat more (gross) variation in task completion in \texttt{tidy kitchen}. This results from the lack of context that was outlined in the body of the paper. For example, for the ``mug on the counter," the agent cannot directly perceive whether the mug is dirty or clean. Verified goals from the agent that the mug should go into the sink or cupboard are selected (i.e., by the Selection process) somewhat arbitrarily (i.e., the system lacks the context that ``dishes out of their storage location should be assumed to be dirty". Because the desired state for this object is always the sink or dishwasher, the agent sometimes places it in the desired location and sometimes not. Collectively, this lack of context accounts for the majority of differences observed for \texttt{tidy kitchen} task completion rate.

\end{document}